\PassOptionsToPackage{table,xcdraw,dvipsnames}{xcolor}
\documentclass{vislang_lab_2026}

\usepackage{pdflscape}
\usepackage{multirow}
\usepackage{xspace}
\usepackage{subcaption}
\usepackage{wrapfig}
\usepackage{siunitx}
\usepackage{listings}
\usepackage[ruled]{algorithm2e}
\usepackage[authoryear,round]{natbib}

\makeatletter
\def\onedot{\futurelet\@let@token\@onedot}
\def\@onedot{\ifx\@let@token.\else.\fi\xspace}
\makeatother

\newcommand{\ie}{\emph{i.e}\onedot}

\newcommand{\MODEL}{{GViT}\xspace}
\newcommand{\custompara}[1]{{\noindent\textbf{#1}\xspace}}
\newcommand{\algofontsize}{6.3pt}

\title{GViT: Representing Images as Gaussians for Visual Recognition}

\author[1]{Jefferson Hernandez}
\author[1]{Ruozhen He}
\author[1]{Guha Balakrishnan}
\author[2]{Alexander C. Berg}
\author[1]{Vicente Ordonez}

\affil[1]{Rice University}
\affil[2]{University of California, Irvine}
\correspondingauthor{jefehern@rice.edu}

\hypersetup{
  pdftitle={GViT: Representing Images as Gaussians for Visual Recognition},
  pdfauthor={Jefferson Hernandez, Ruozhen He, Guha Balakrishnan, Alexander C. Berg, Vicente Ordonez}
}

\begin{document}

\maketitle

\begin{abstract}
We introduce \MODEL, a supervised classification framework that uses patch-conditioned 2D Gaussian primitives as input tokens for ViT backbones. A ViT-S-sized encoder predicts each primitive's position, scale, orientation, color, and opacity in one forward pass. The same parameters are constrained to reconstruct the image through a differentiable renderer, as well as serve as classifier inputs. During training, classifier-gradient guidance shapes primitive geometry and appearance under an image reconstruction constraint. On ImageNet-1K, \MODEL-B achieves 76.9\% top-1 accuracy, compared with 79.8\% for ViT-B/16 trained with the same recipe. Gaussian tokens also provide explicit spatial support for explanation maps. Grad-CAM over Gaussians improves on patch-based Grad-CAM in segmentation and token-deletion probes, while other probe comparisons are comparable to a ViT. A covariance-based foreground diagnostic on the PASCAL VOC segmentation task further measures the geometry learned through supervised guidance. These results establish the feasibility of recognition from renderable Gaussian parameters and characterize how task supervision shapes this inspectable representation.
\end{abstract}

% \vspace{-0.2in}
\section{Introduction}
\label{sec:intro}
\vspace{-0.1in}

Computer vision has long treated the pixel grid~\citep{kirsch1998seac} as its default interface.  Convolutional networks~\citep{NIPS2012_4824, DBLP:journals/corr/SimonyanZ14a, he2016residual, Liu_2022_CVPR}, Vision Transformers (ViTs)~\citep{dosovitskiy2020image, steiner2022how, dehghani2023scaling}, and self-supervised models such as masked autoencoders~\citep{he2022masked} and DINO~\citep{oquab2024dinov} all build on pixels or fixed patches---convenient and effective, but only weakly tied to the continuous geometry of natural images~\citep{darcet2024vision}. We ask whether a classifier can instead consume a compact set of explicit geometric primitives.

\begin{figure}[t]
  \centering
  \includegraphics[width=\textwidth]{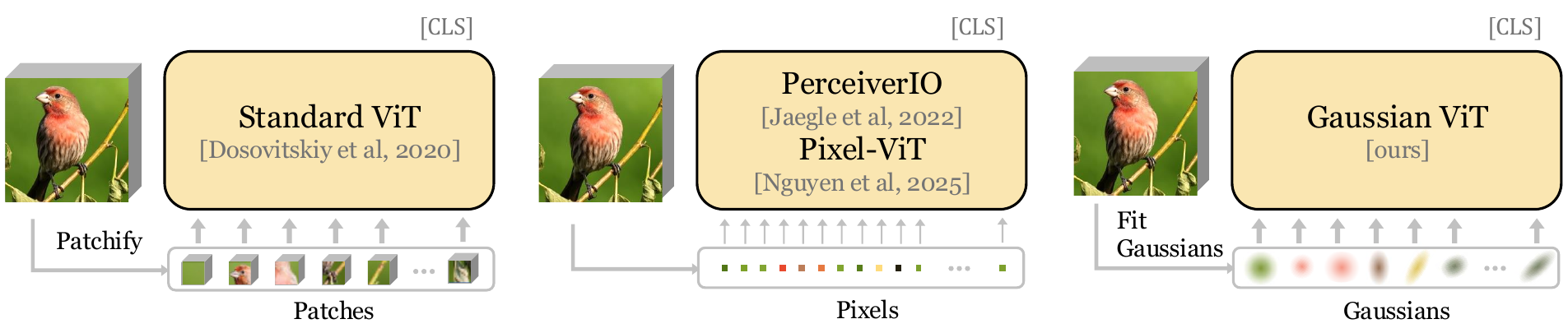}
  \caption{\MODEL replaces fixed patch tokens with a compact set of predicted 2D Gaussians. The encoder fits these primitives in one forward pass, making Gaussian tokens practical as the input interface of a ViT classifier.}
  \label{fig:lead}
\vspace{-0.2in}
\end{figure}

We introduce \MODEL (Fig.~\ref{fig:lead}), a Gaussian Vision Transformer that represents each image as a set of 2D Gaussian primitives and feeds their parameters to a ViT classifier. A ViT-S-sized encoder predicts each Gaussian's position, covariance, color, and opacity in one forward pass. These parameters define a smooth spatial footprint and rendered appearance. The primitives can move, change extent, and overlap to allocate image detail. A fixed differentiable renderer reconstructs the image from the same parameters that the classifier consumes, without a learned image decoder. Classifier gradients guide this representation toward class-relevant regions while reconstruction constrains its appearance. We study this joint design of geometric tokens and task-driven guidance.

This places \MODEL among alternatives to rigid patch grids, including superpixels~\citep{ren2003learning, achanta2012slic, jampani2018superpixel}, slot-based models~\citep{locatello2020object,pmlr-v202-biza23a, fan2024adaptive}, byte- or pixel-level interfaces~\citep{horton2023bytes,nguyen2024image}, and learned visual tokens~\citep{bao2021beit, Yu2021VQGAN, Mao2022DAT}. Our distinction is to use continuous, differentiably rendered Gaussians as the classifier tokens themselves, yielding a compact mid-level canvas for recognition and explanation.

On ImageNet-1K, \MODEL-B reaches 76.9\% top-1 accuracy, 2.9 points below ViT-B/16 trained with the same recipe. The full system combines a 22M-parameter encoder with an 86M-parameter classifier and uses 39.2 GMAC at 224\(\times\)224, compared with 17.6 GMAC for ViT-B/16. These measurements make the accuracy and cost of the Gaussian interface explicit. Our contributions are patch-conditioned Gaussian tokenization, classifier-gradient guidance under reconstruction, and quantitative probes of the learned representation. The explanation probes measure segmentation overlap and the effect of deleting classifier input tokens. The covariance diagnostic measures whether supervised guidance makes foreground structure visible in Gaussian geometry. Together, these experiments test recognition and inspectability through an explicit geometric interface.

\section{Related Work}
% \vspace{-0.1in}
\label{sec:related}
\vspace{-0.1in}

\custompara{Structured and adaptive tokenization.}
GaussianImage~\citep{zhang2024gaussianimage} and Gaussian Masked Autoencoders (GMAE)~\citep{rajasegaran2025gaussianmae} also use 2D Gaussians, but for compression or reconstruction; GMAE discards them at inference. \MODEL instead keeps Gaussians as classifier tokens and trains their placement with classifier guidance. Other work departs from patch grids via image-dependent regions: SuiT~\citep{lew2024superpixel} uses superpixels, DART~\citep{yin2025dart} learns adaptive partitions, and DPT~\citep{chen2021dpt} samples deformable patches. These are complementary: superpixels and adaptive patches define discrete regions, while Gaussian tokens are continuous primitives whose parameters can be differentiated and steered by the recognition loss. Related structured-token work includes learnable superpixels~\citep{jampani2018superpixel}, object-centric slots~\citep{locatello2020object,fan2024adaptive}, dense part discovery~\citep{ziegler2022self}, and geometry-aware primitives such as multi-plane images~\citep{tucker2020single}, but \MODEL provides a supervised classifier interface rather than only reconstruction, grouping, or scene representation.

\custompara{Vision beyond pixels and patches.} 
Other interfaces avoid raw RGB patches without introducing explicit primitives: JPEG/DCT tokens, raw bytes, next-byte prediction, and individual pixels~\citep{park2023rgb,horton2023bytes,wu2024beyond,nguyen2024image}. Learned-tokenizer methods instead use dVAE, vector-quantized, reconstruction, or self-supervised patch targets~\citep{bao2021beit,Yu2021VQGAN,he2022masked,oquab2024dinov}. These methods show that recognition need not consume raw RGB patches, but their tokens are bytes, pixels, discrete codes, or latent patch features rather than an image-specific primitive canvas. \MODEL remains patch-conditioned in the encoder but exposes compact 2D Gaussian tokens to the classifier, making support, appearance, and task-driven relocation directly inspectable.

\section{Method}
\label{sec:method}
\vspace{-0.1in}

\begin{figure*}[t]
    \centering
    \includegraphics[width=0.9\linewidth]{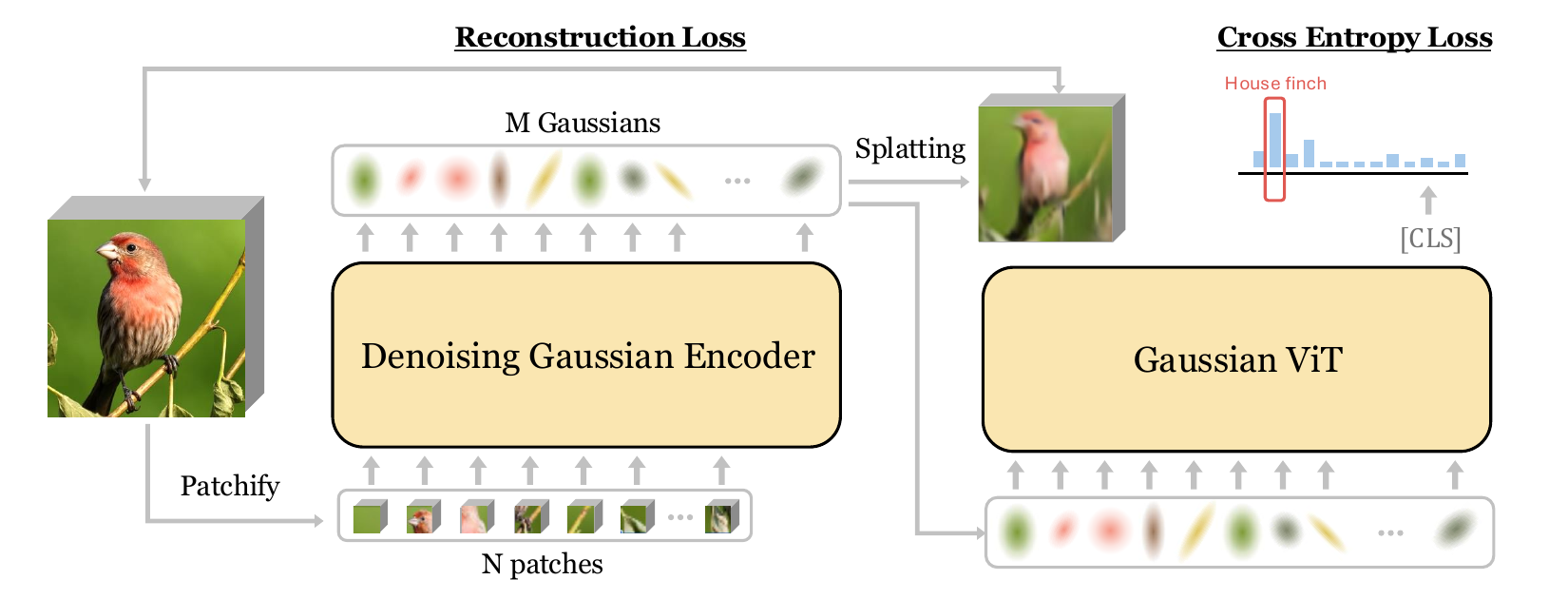}
    % \vspace{-.1in}
    \caption{Overview of \MODEL. A Denoising Gaussian Encoder predicts 2D Gaussians from image patches. The Gaussians are rendered for reconstruction and passed as tokens to a ViT classifier. Classifier-gradient guidance shapes the primitives during training. Algorithm~\ref{alg:method} gives the training procedure.}
    \label{fig:system-figure}
    \vspace{-0.2in}
\end{figure*}

We encode each image as a compact set of 2D Gaussians and optimize their predicted properties for recognition (Fig.~\ref{fig:system-figure}). The Denoising Gaussian Encoder processes image patch embeddings together with $K$ randomly sampled Gaussian inputs. It predicts the parameters of $K$ primitives. A differentiable renderer reconstructs the image from these parameters, and a ViT classifier~\citep{dosovitskiy2020image} reads features built from the same parameters. During guidance steps, classifier gradients update only the encoder under reconstruction regularization (Sec.~\ref{sec:guidance}).

Guidance and Gaussian tokenization form a joint design. Reconstruction and recognition can favor different primitive layouts: background detail can improve reconstruction without helping classification. Classifier guidance supplies a task-alignment signal, and reconstruction constrains the rendered image. The no-guidance setting measures this representation without the additional guidance updates. Reconstruction serves as a fitting constraint during supervised training; MAE~\citep{he2022masked} instead uses reconstruction as a self-supervised pretext.

% \vspace{-0.15in}
\subsection{Gaussian primitives and rendering}
\vspace{-0.1in}

Our representation follows the spirit of optimization-based 3D Gaussian splatting~\citep{kerbl20233d}, adapted to 2D. An image is represented by a set \(\{\mathbf g_j\}_{j=1}^{K}\) of 2D Gaussian primitives. Each Gaussian has center \(\boldsymbol\mu\in\mathbb R^{2}\), covariance \(\mathbf\Sigma\in\mathbb R^{2\times 2}\), color \(\boldsymbol\rho\in\mathbb R^{3}\), and opacity \(o\in\mathbb R\).

Rendering is performed by orthographic volumetric splatting with no camera transform: all Gaussians lie on a common \(z\)-plane and are composited with standard alpha blending. This setup yields a differentiable renderer so that image-space losses can backpropagate to all Gaussian attributes.

We parameterize the covariance by scale and rotation:
\[
\mathbf\Sigma \;=\; \mathbf R(\alpha)\,\mathbf S\,\mathbf S^\top\,\mathbf R(\alpha)^\top,
\qquad
\mathbf S \;=\; \operatorname{diag}\!\bigl(\tilde{\boldsymbol{\sigma}}\bigr),\quad
\tilde{\boldsymbol{\sigma}}\in\mathbb R_{+}^{2},
\]
where \(\alpha\) is the in-plane rotation angle and \(\tilde{\boldsymbol{\sigma}}=(\tilde{\sigma}_x,\tilde{\sigma}_y)\) are the pre-rotation standard deviations along the horizontal and vertical axes. We produce positive scales from unconstrained \emph{scale logits} \(\mathbf s\in\mathbb R^{2}\) using a bounded sigmoid:
\[
\tilde{\boldsymbol{\sigma}} \;=\; c \cdot \sigma(\mathbf s)
\quad\text{with}\quad
\sigma(u)=\frac{1}{1+e^{-u}}, \; c>0 \text{ fixed.}
\]
A single Gaussian primitive is thus described by vector
\(
\mathbf g \;=\; \{\boldsymbol\mu,\mathbf s,\alpha,\boldsymbol\rho,o\}\in\mathbb R^{9}.
\)
Following \citet{rajasegaran2025gaussianmae}, the cap \(c\) controls the maximum spatial extent and prevents Gaussians from degenerately covering the entire image.

\subsection{Denoising Gaussian Encoder}
\vspace{-0.1in}

Given an input image, we split it into $N$ patches and project them to $N\times d_{\text{enc}}$ embeddings. A separate linear projection maps $K$ randomly sampled Gaussian inputs $\mathbf g_0\in\mathbb R^{K\times9}$ to $K\times d_{\text{enc}}$ embeddings. We concatenate both sequences and apply self-attention over $(N+K)\times d_{\text{enc}}$ activations. The random Gaussian inputs serve as queries whose output latents predict the primitives. The learned-query ablation instead uses learned inputs. We call the main encoder \emph{denoising} because it refines random Gaussian initializations in one forward pass. Patch conditioning is an implementation choice that replaces per-image Gaussian fitting with a feed-forward prediction. The classifier receives only features from the predicted Gaussian parameters; their centers, scales, and rotations are free to vary independently of the patch grid. Pixel or convolutional conditioning is a possible extension that we do not evaluate here.

The encoder processes the concatenated token sequence and outputs latents \(\mathbf x_i\in\mathbb R^{d_{\text{enc}}}\) for \(i=1,\dots,N+K\). We discard the \(N\) patch latents and map the \(K\) Gaussian latents through an MLP to predict residuals:
\[
\Delta\mathbf g \;=\; \mathrm{MLP}\bigl(\mathbf x_{N+1:N+K}\bigr),
\qquad
\hat{\mathbf g}\;=\;\mathbf g_0+\Delta\mathbf g.
\]
The predicted Gaussians \(\hat{\mathbf g}\) are rendered under the fixed orthographic camera to obtain a reconstruction. The encoder is trained with a pixel reconstruction term \(\mathcal L_{\text{pix}}\) (BCE on normalized RGB by default; MSE is evaluated in ablations) and a perceptual term \(\mathcal L_{\text{perc}}\) (D-SSIM~\citep{wang2004image}), combined as $\mathcal L_{\text{rec}} \;=\; \mathcal L_{\text{pix}} + \lambda_{\text{perc}}\,\mathcal L_{\text{perc}}$.
%\[
%\mathcal L_{\text{rec}} \;=\; \mathcal L_{\text{pix}} + \lambda_{\text{perc}}\,\mathcal %L_{\text{perc}}.
%\]
%In practice, the initial Gaussian set \(\mathbf g_0\) can be drawn from simple unsupervised segmentations (e.g., \(k\)-means in color space, backpropagation-based segmentation~\citep{kanezaki2018unsupervised}, or Felzenszwalb–Huttenlocher~\citep{felzenszwalb2004efficient}), akin to point-cloud initializations used in 3D Gaussian splatting. 
The reconstruction and classification objectives optimize the encoder through the predicted Gaussian parameters. The downstream classifier uses centers $\boldsymbol\mu$, log-scales, $\sin$/$\cos$ of the angle $\alpha$, color $\boldsymbol\rho$, and opacity $o$. This feature construction maps each 9-parameter primitive to 10 features, which a linear projection maps to $d_{\text{cls}}$ dimensions. The resulting $K\times d_{\text{cls}}$ sequence exposes both appearance and geometry to the classifier.

% \vspace{-0.15in}
\subsection{Classifier-steered token relocation}
\label{sec:guidance}
\vspace{-0.1in}

Adversarial attacks typically \emph{reduce} a model's confidence by perturbing inputs in the direction of \(\nabla_x \mathcal H(f(x;\psi),y)\), where \(\mathcal H\) is the cross-entropy and \(f(\cdot;\psi)\) is a classifier with parameters \(\psi\). We adopt the opposite perspective: we use classifier gradients \emph{constructively} to relocate Gaussians toward regions that support the correct class, while a reconstruction term keeps the renderings faithful. This guidance step is the intended use of the continuous Gaussian parameterization: it supplies the task-aligned relocation signal that patch-conditioned reconstruction alone does not provide.

% \vspace{0.04in}
\noindent\textbf{Setup and notation.}
Let \(\mathbf G=\{\hat{\mathbf g}_j\}_{j=1}^{K}\) collect all predicted Gaussian parameters, produced by an encoder with parameters \(\phi\), and let \(f(\cdot;\psi)\) denote the classifier that takes Gaussian tokens as input. With ground-truth label \(y\), the classification loss is
\[
\mathcal L_{\text{cls}} \;=\; \mathcal H\!\bigl(f(\mathbf G;\psi),\,y\bigr),
\]
and the reconstruction loss \(\mathcal L_{\text{rec}}\) is as defined above. Guidance gradients pass through the predicted Gaussian parameters to the encoder weights $\phi$. The classification branch reads these parameters directly, while the reconstruction branch passes through the renderer. Guidance updates the encoder weights; the predicted Gaussians are the representation being steered.

\begin{figure}[t!]
  \centering
  \small
\begin{subfigure}[b]{0.24\textwidth}
    \includegraphics[width=\linewidth]{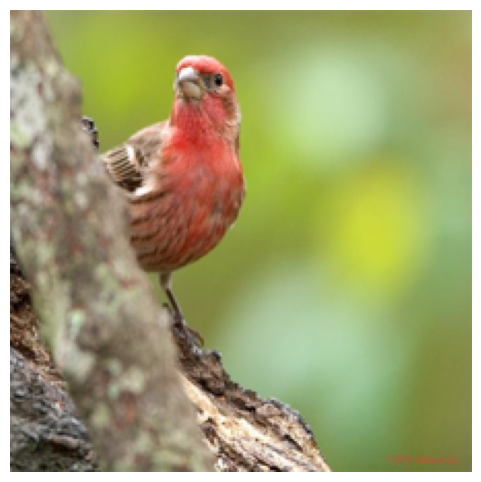}
    \subcaption{Original Image}
    \label{fig:bird_sgd1}
\end{subfigure}\hfill
  \begin{subfigure}[b]{0.24\textwidth}
    \includegraphics[width=\linewidth]{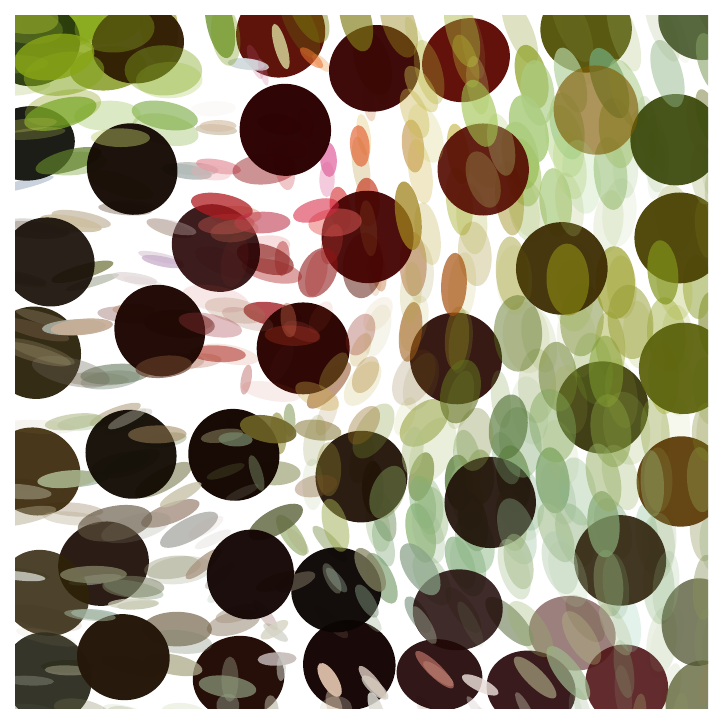}
    \subcaption{SGD Gaussians}
    \label{fig:bird_sgd2}
  \end{subfigure}\hfill
  \begin{subfigure}[b]{0.24\textwidth}
    \includegraphics[width=\linewidth]{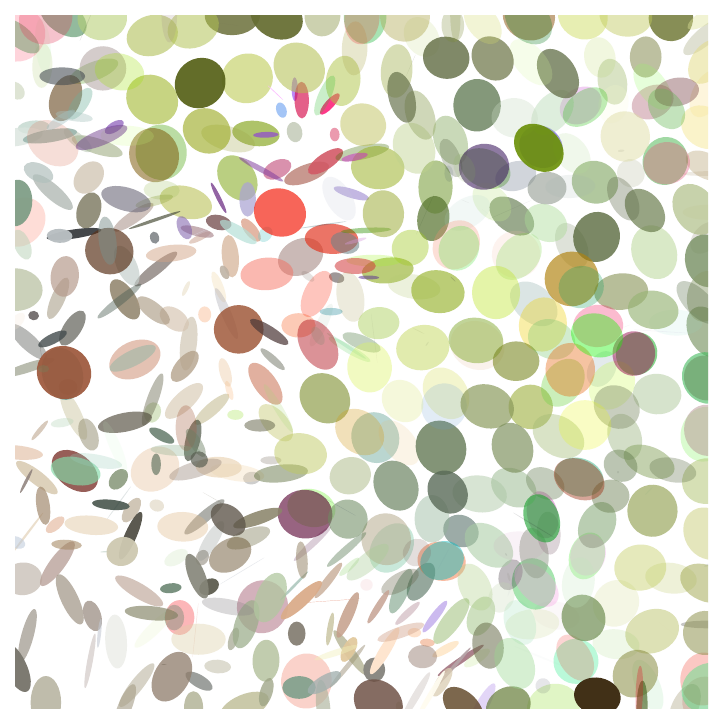}
    \subcaption{No Guidance}
    \label{fig:bird_wo_guid}
  \end{subfigure}\hfill
  \begin{subfigure}[b]{0.24\textwidth}
    \includegraphics[width=\linewidth]{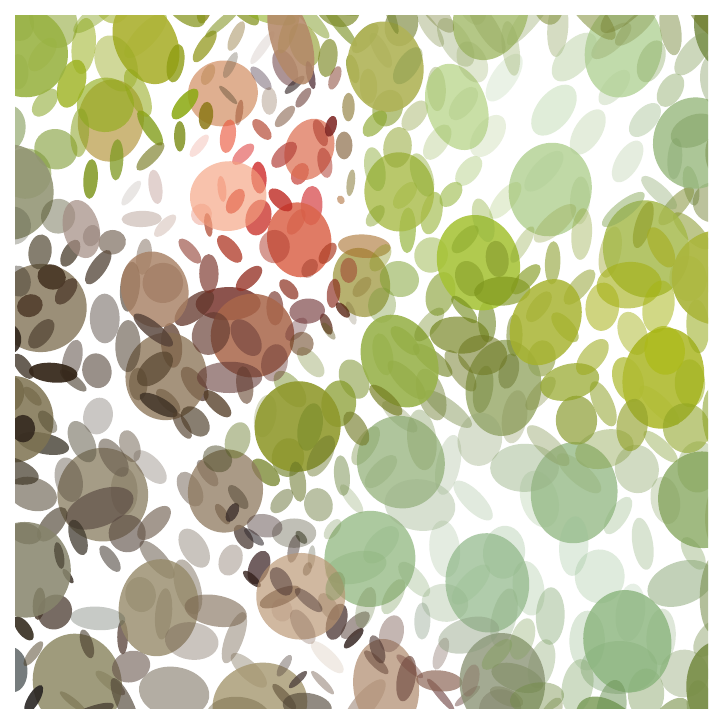}
    \subcaption{Gradient Guidance}
    \label{fig:bird_w_guid}
  \end{subfigure}
  \caption{{Visualization of Gaussians for an image under three regimes.} (a) Original image, (b) Gaussians fitted via SGD reconstruction-only, (c) End-to-end predicted Gaussians without classifier guidance, and (d) Gaussians steered by classifier-gradient guidance. We use an ellipse sized to 2 standard deviations for each Gaussian to visualize.}
  \label{fig:bird_splat_comparison}
  \vspace{-0.2in}
\end{figure}

\custompara{Encoder-only classifier-gradient relocation.}
During guidance steps, we freeze classifier parameters \(\psi\) and backpropagate only through the encoder weights \(\phi\). We define \(\widetilde{\nabla}_{\phi}\) as the \emph{gradient passed to a minimization optimizer} (\ie the optimizer subtracts \(\eta\,\widetilde{\nabla}_{\phi}\) from \(\phi\)) and combine reconstruction and classification gradients additively:
\begin{equation}
\label{eq:guidance_update}
\widetilde{\nabla}_{\phi}
\;=\;
\nabla_{\phi}\mathcal L_{\text{rec}}
\;+\;
\gamma\,\nabla_{\phi}\mathcal L_{\text{cls}},
\qquad
\gamma\in[0,0.1].
\end{equation}
The second term \(+\gamma\nabla_{\phi}\mathcal L_{\text{cls}}\) is gradient \emph{descent} on \(\mathcal L_{\text{cls}}\) routed through the encoder: subtracting it during the optimizer step \emph{pulls} Gaussians toward class-salient regions, while \(\nabla_{\phi}\mathcal L_{\text{rec}}\) acts as a reconstruction regularizer that keeps the layout image-faithful and prevents degenerate configurations.
Appendix~\ref{app:guidance_intrinsic} gives local-linear intuition for guidance in Gaussian parameter space. The gradient argument also applies to other differentiable continuous representations; the explicit geometric meaning of the updated parameters is the property we study here.

% \vspace{0.02in}
\noindent{\bf Three-phase training schedule.}
We alternate reconstruction and recognition so that classifier gradients are informative before they are used for guidance:
\begin{enumerate}[label=(\roman*), leftmargin=2.2em]
\item \textbf{Reconstruction warm-up.} Train the Denoising Gaussian Encoder alone, minimizing \(\mathcal L_{\text{rec}}\) (e.g., start with \(\mathcal L_{\text{pix}}\) and add \(\lambda_{\text{perc}}\mathcal L_{\text{perc}}\) after a short burn-in). This produces image-faithful Gaussians before any task signal is involved.
\item \textbf{Classifier warm-up.} Freeze the encoder \(\phi\) and train the classifier \(f(\cdot;\psi)\) on the current Gaussians with \(\mathcal L_{\text{cls}}\), so that its gradients carry semantic signal before they are reused as guidance.
\item \textbf{Joint guided training.} Unfreeze all parameters and update \(\phi\) with the composite gradient \eqref{eq:guidance_update}, interleaving guidance steps with standard gradient updates (e.g., one guidance step after every \(m\) normal steps). The guidance weight \(\gamma\) is linearly ramped to a small maximum (e.g., \(\gamma_{\max}=0.1\)) for stability.
\end{enumerate}

% \vspace{-0.15in}
\section{Experiment settings}
\label{sec:exp_settings}
% \vspace{-0.15in}

We evaluate on ImageNet-1K~\citep{russakovsky2014imagenet} and seven transfer datasets. Ablations use the Mini-ImageNet split from \citet{vinyals2016matching}. GViT training spans 400 epochs across reconstruction warm-up, classifier warm-up, and joint training. We use AdamW~\citep{loshchilov2017decoupled}, a base learning rate of $10^{-4}$, and cosine learning-rate decay. Our patch-based ViT baselines use the same classification recipe; reconstruction and guidance are specific to GViT. We use PyTorch with bfloat16 mixed precision. Full GViT ImageNet training takes about 12 hours on eight NVIDIA A100 GPUs with 40 GB each, or about 96 GPU-hours. Appendix~\ref{supp:impl} gives implementation and training settings. We report model-forward inference GMAC separately from training time and end-to-end throughput.

\begin{table}[t]
% \vspace{-0.1in}
  \newcommand{\grayfont}{\color{black!60}}
  \centering
  % \footnotesize % Stepped down from \small to fit LNCS limits better
  \renewcommand{\arraystretch}{1.5}
  \setlength{\tabcolsep}{1.5pt} % Slightly tighter spacing
    \caption{Top-1 classification accuracies (\%) on ImageNet-1K (IN-1K) and seven transfer datasets for Gaussian ViT variants and literature ViT baselines. C10, C100, SUN, and Flwr denote CIFAR-10, CIFAR-100, SUN397, and Flowers. AVG is the mean of the positive values in the seven transfer columns, excluding IN-1K and missing entries; dashes denote unreported results. Literature baselines use their source protocols~\citep{dosovitskiy2020image,steiner2022how}.}
  \label{tab:main_result}
  \begin{tabular}{l l c c c c c c c c c}
    \toprule
     \multirow{2}{*}{\textbf{Model}} & 
     \multirow{2}{*}{\textbf{Input}} &
     & \multicolumn{8}{c}{\textbf{Fine-grained Benchmarks}} \\
    \cmidrule(lr){4-11}
     & & \textbf{IN-1K} & \textbf{AVG} & \textbf{C10} & \textbf{C100} & \textbf{CUB} & \textbf{SUN} & \textbf{Flwr} & \textbf{Pets} & \textbf{DTD} \\
    \midrule
    \grayfont ViT-B/32~\citep{dosovitskiy2020image}  & Patches & \grayfont 71.4 & \grayfont 85.6 &\grayfont -    &\grayfont 87.6 &\grayfont 92.6 &\grayfont 73.8 &\grayfont 94.4 &\grayfont 92.2 &\grayfont 72.7 \\
    \grayfont ViT-B/16~\citep{dosovitskiy2020image} & Patches  &\grayfont 77.9 & \grayfont 92.1 &\grayfont 98.1 &\grayfont 87.1 &\grayfont - &\grayfont - &\grayfont 89.5 &\grayfont 93.8 &\grayfont - \\
    \grayfont ViT-B/16~\citep{steiner2022how}  & Patches  &\grayfont 78.7 & \grayfont 89.0 &\grayfont 98.1 &\grayfont 87.8 &\grayfont 93.0 &\grayfont 75.3 &\grayfont 97.6 &\grayfont 93.2 &\grayfont 77.8 \\
   \midrule
    \MODEL-B {(no guid)} & {Gaussians}   & 73.6 & 84.9 & 96.6 & 85.7 & 88.3 & 67.1 & 92.3 & 92.2 & 71.9 \\
    \rowcolor{blue!5}
    \MODEL-B {(w/guid)}  & {Gaussians}  & {76.9} & 87.1 & {98.6} & {87.7} & {91.2} & {68.3} & {96.0} & {94.5} & {73.4} \\
    \bottomrule
  \end{tabular}
  \vspace{-0.2in}
\end{table}

\custompara{Architecture and cost.} The Gaussian encoder has 22M parameters, similar in size to ViT-S. It uses either learned queries or random Gaussian inputs with residual prediction, as specified in each experiment.

At $224\times224$, ViT-B/16 trained with our recipe uses 86M parameters, 196 patch tokens, and 17.6 GMAC per forward pass. DeiT-B/16 uses 86.57M parameters, 196 patch tokens, and 17.56 GMAC. Both GViT-B variants use a 22M-parameter encoder and an 86M-parameter classifier (108M total), process $196+512$ encoder tokens and 512 classifier tokens, and require 39.2 GMAC.

Token counts exclude CLS. We use fvcore's convention of one count per multiply-accumulate; one GMAC is $10^9$ such operations. These counts describe model inference, separately from image loading and decoding; the full GViT-S forward pass requires 24.2 GMAC. On one A100-40GB, the end-to-end encoded-image pipeline, including disk decoding, processes 255.3 images/s for GViT-B at $224\times224$ with 2.9 GB peak memory. 

% \vspace{-0.15in}
\section{Results and Ablations}
\label{sec:results}
\vspace{-0.1in}

Our experiments compare \MODEL against standard patch-based ViTs in large-scale classification (Sec.~\ref{sec:main-results}) and interpretability (Sec.~\ref{sec:interpretability}). We also analyze the impact of our design choices %, such as the number and scale of Gaussians 
(Sec.~\ref{sec:ablations}); additional diagnostics for representational shortcuts are deferred to Appendix~\ref{sec:leakage}.

% \vspace{-0.1in}
\subsection{Main result}
\label{sec:main-results}
\vspace{-0.1in}
With guidance, \MODEL-B reaches 76.9\% top-1 on ImageNet-1K (Table~\ref{tab:main_result}). As a matched-recipe reference, ViT-B/16 reaches 79.8\%. GViT-B uses 108M parameters and 39.2 GMAC, compared with about 86M and 17.6 GMAC for ViT-B/16. The no-guidance variant reaches 73.6\%. This 3.3-point difference measures the effect of the guidance procedure, which adds encoder updates; the update-count control in Sec.~\ref{sec:ablations} tests the classifier-gradient term separately. The 78.7\% literature result from \citet{steiner2022how} uses its own training and fine-tuning protocol. Nondistilled DeiT-B/16 reaches 81.8\%~\citep{pmlr-v139-touvron21a}. Appendix~\ref{app:extra_results} reports capacity, sequence lengths, cost, and training settings for alternative token interfaces. 

We also evaluate transfer to seven specialized benchmarks by freezing the Gaussian \emph{generator} %learned on ImageNet-1K 
and fine-tuning the ViT classifier (backbone and head) with layer-wise learning-rate decay. Guidance improves accuracy on every displayed transfer dataset. The unweighted mean over these seven tasks rises from 84.9\% to 87.1\%. Flowers gains 3.7 points. These results show that the guided representation retains useful information when the Gaussian encoder is frozen for transfer. 

\begin{table}[t!] % Use table* for a wider, two-column-spanning table
\centering
% \footnotesize
\newcommand{\lightfont}{\color{black!60}}
\renewcommand{\arraystretch}{1.1}
\setlength{\tabcolsep}{4pt} % Adjust column spacing as needed
\caption{
    Explainability performance on ImageNet-Seg and ImageNet-Val under perturbations.
    %We compare different explainability methods.
    Perturbation deletes classifier input tokens on a 10--70\% schedule. ViT uses black input patches. GViT uses a position-only null: the center remains, covariance is fixed and isotropic, and color and opacity are zero.
    %Note that ViT requires different methods (Rollout for segmentation, CDAM for perturbation) to achieve its best results.
    %\textbf{GViT achieves top segmentation performance with Grad-CAM} and remains competitive on perturbation under the matched protocol, while also performing reasonably well with our simple heuristic \(\det(\Sigma)\). 
    ViT segmentation numbers are from \citet{chefer2021transformer}; perturbation results use the stated token-deletion protocol. Bold values mark the best probe within each model for each metric.
    %\vspace{-.05in}
    }
\label{tab:explainability_combined}
\begin{tabular}{l l c c c c c}
\toprule
& & \multicolumn{3}{c}{\textbf{ImageNet-Seg}} & \multicolumn{2}{c}{\textbf{Perturbation}} \\
\cmidrule(lr){3-5} \cmidrule(lr){6-7}
\textbf{Method} & \textbf{Model} & \textbf{Pixel Acc} & \textbf{mIoU} & \textbf{mAP} & \textbf{Neg (\(\uparrow\))} & \textbf{Pos (\(\downarrow\))} \\

\midrule
\rowcolor{blue!5}
 & ViT & -- & -- & -- & -- & -- \\
\rowcolor{blue!5}
\multirow{-2}{*}{\(\det(\Sigma)\)} & GViT & 55.45 & 38.65 & 66.75 & 36.4 & 41.3 \\

\midrule
\multirow{2}{*}{Raw Attn} & ViT & 67.84 & 46.37 & 80.24 & 50.00 & 23.70 \\
& GViT & 61.09 & 42.89 & 75.68 & 42.0 & 29.9 \\

% \midrule
% \multirow{2}{*}{Rollout {\lightfont \citep{abnar2020quantifying}}} & ViT & 73.54 & 55.42 & 84.76 & 51.30 & 20.40 \\
% & GViT & 68.31 & 49.94 & 81.76 & 49.30 & 22.60 \\

\midrule
\multirow{2}{*}{Rollout} & ViT & \textbf{73.54} & \textbf{55.42} & \textbf{84.76} & 51.3 & 20.4 \\
& GViT & 68.31 & 49.94 & 81.76 & 49.0 & 23.0 \\

% \midrule
% \multirow{2}{*}{CDAM {\lightfont \citep{brocki2023class}}} & ViT & 68.80 & 48.50 & 81.50 & 54.80 & 17.30 \\
% & GViT & 70.25 & 50.86 & 82.31 & 50.23 & 21.58 \\

\midrule
\multirow{2}{*}{CDAM} & ViT & 68.80 & 48.50 & 81.50 & \textbf{54.8} & \textbf{17.3} \\
& GViT & 70.25 & 50.86 & 82.31 & 51.7 & 20.5 \\

\midrule
\rowcolor{blue!5}
 & ViT  & 64.44 & 40.82 & 71.60 & 46.80 & 27.70 \\
\rowcolor{blue!5}
\multirow{-2}{*}{Grad-CAM}& GViT & \textbf{74.31} & \textbf{54.94} & \textbf{84.67} & \textbf{53.9} & \textbf{17.8} \\
\bottomrule
\end{tabular}
\end{table}

\subsection{Explainability of Gaussian Representations.}
\label{sec:interpretability}
\vspace{-0.1in}

\custompara{Setup.}
Given a token sequence \([\mathrm{CLS}], g_1,\dots,g_K\), let \(A^{(l)}\) be the head-averaged attention maps and \(h^{(l)}\) the hidden states for block \(l\). All methods below produce a token-level relevance score \(r\in\mathbb{R}^{K+1}\). For visualization, we drop the \([\mathrm{CLS}]\) entry and rasterize \(r_{1:K}\) to pixels via each token's Gaussian footprint.

\custompara{Gaussian-to-Image Projection.}
All methods project token-level relevance \(r_i\) into a heatmap \(H(x,y)\) using a normalized, weighted sum of the tokens Gaussian spatial extents:
\begin{equation}
H(x,y)=\frac{\sum_{i=1}^{K} r_i\,M_i(x,y)}{\sum_{i=1}^{K} M_i(x,y)+\varepsilon}.
\label{eq:gauss2img}
\end{equation}
Here, \(M_i(x,y)\) is the soft mask for token \(i\), defined by its 2D Gaussian probability density. In practice, we rasterize \(M_i\), truncate at 3-sigma, and \(\mathcal{L}_1\)-normalize its discrete samples before applying \eqref{eq:gauss2img}.

\begin{figure*}[t]
  \centering
  \includegraphics[width=\linewidth]{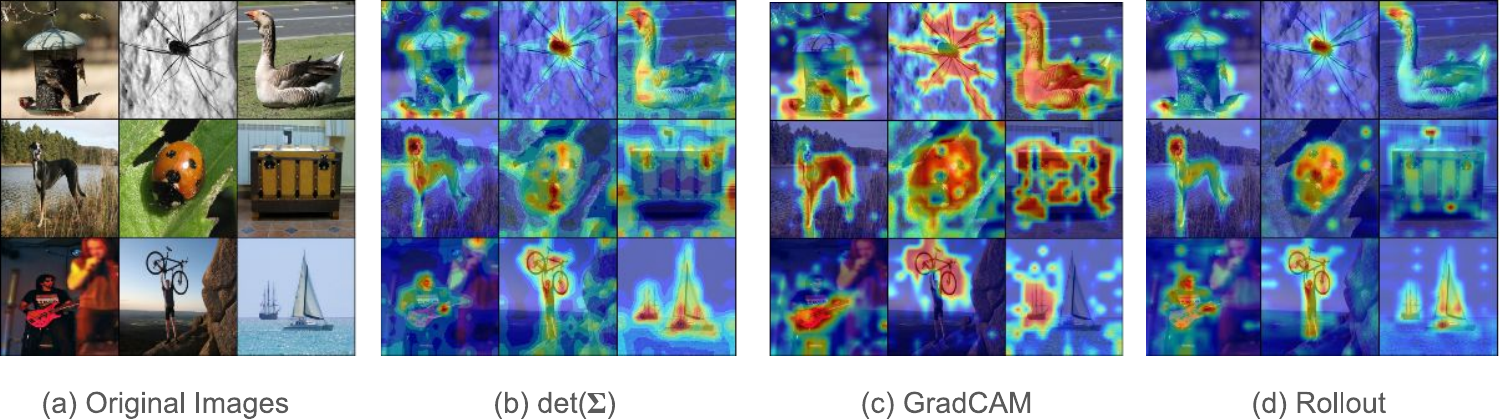}
  \caption{Gradient‐guided Gaussians exhibit a natural interpretability component. (a) Original input images. (b) Determinant of the covariance (\(\det(\Sigma)\)), which clusters Gaussian primitives in class‐salient regions. (c) GradCam over the learned Gaussians. (d) Rollout of the attentions.}
  \label{fig:interpretability}
  \vspace{-.3in}
\end{figure*}

\custompara{Explainability Methods.}
{\em (1) Raw attention (class-agnostic). }
We use the head-averaged attention from the final block \(L\). We set the relevance vector \(r{=}\bar A^{(L)}_{\mathrm{CLS},:}\) and project using \eqref{eq:gauss2img}. {\em (2) Attention rollout (class-agnostic, global).}
We compute the attention rollout~\citep{abnar2020quantifying} to aggregate contributions across layers. Let \(\tilde A^{(l)}{=}I{+}\bar A^{(l)}\). The rollout matrix is
$R{=}\tilde A^{(1)}\tilde A^{(2)}\cdots\tilde A^{(L)}.$
We set \(r{=}R_{\mathrm{CLS},:}\) and project with \eqref{eq:gauss2img}.  {\em (3) Grad-CAM~\citep{selvaraju2017grad} (class-specific).}
For a target logit \(y_c\), we backpropagate to the final hidden states \(h^{(L)}\) to get gradients \(D{=}\partial y_c/\partial h^{(L)}\). The token weights are the channel-wise average gradients, \(a_i{=}\tfrac{1}{d}\sum_{k} D_{i,k}\).  The token relevances \(r_i{=}\mathrm{ReLU}(\sum_k a_i\,h^{(L)}_{i,k})\)
which is projected using \eqref{eq:gauss2img} to produce a class-specific map. {\em (4) CDAM on Gaussians (class-specific).}
Following \citet{brocki2023class}, we compute \(S_{i,c}{=}\sum_j g_{i,j}\,\frac{\partial f_c}{\partial g_{i,j}}\) over the token input vector \(g_i\) to the final attention layer. We set \(r_i{=}S_{i,c}\) and, unlike the patch-grid projection in prior work, we render \(r\) with \eqref{eq:gauss2img}, leveraging each token continuous footprint. \textbf{(5) Geometric prior (ours): \(\det(\Sigma)\).}
As a class-agnostic baseline, we use the negative log-determinant \(r_i{=}-\log\det(\Sigma_i)\) as a saliency prior. This assumes that small-volume Gaussians 
% (high \(r_i\)) 
cluster on discriminative regions.

\custompara{Why these probes are appropriate.}
These probes cover attention, gradients, and geometry, with both class-specific and class-agnostic scores. Gaussian spatial support lets each score project onto the same image canvas through \eqref{eq:gauss2img}. This property makes projection explicit, but does not guarantee foreground overlap or explanation faithfulness. Segmentation and token deletion measure these empirical properties of the learned representation.

\custompara{Datasets and Metrics.}
We evaluate on ImageNet-val~\citep{russakovsky2014imagenet} and ImageNet-Seg~\citep{guillaumin2014imagenet}. All probes produce a token-level relevance vector $r$, which we project to pixel space via \eqref{eq:gauss2img} and min--max normalize per image to $[0,1]$ before scoring.
\textbf{Segmentation metrics:} we use heatmaps as soft segmentations and report pixel accuracy (binarized at the per-image mean intensity, following~\citet{chefer2021transformer}), mIoU, and mAP (threshold-agnostic).
\textbf{Perturbation metrics:} we sort tokens by relevance $r_i$ and delete fractions from 10\% to 70\%. \textit{Positive} deletes the most relevant first (lower top-1 AUC is better); \textit{Negative} deletes the least relevant first (higher AUC is better). For ViT, blackening a complete patch replaces its embedding with the linear projection of zeros, including projection bias, and retains its positional embedding. For GViT, the intervention acts on predicted parameters before the classifier projection, without re-encoding the image. The position-only null retains the center, resets covariance to the identity matrix ($\Sigma=I$), and sets color and opacity to zero. Both sequences keep their length. This protocol tests each classifier's input representation. It does not equate the removed image area or test GViT after pixel deletion and re-encoding; the retained Gaussian centers can still carry image information.

\custompara{Takeaways.}
Grad-CAM gives the strongest results among the evaluated GViT probes (Table~\ref{tab:explainability_combined}). It exceeds ViT Grad-CAM on segmentation and token deletion, with Negative/Positive scores of 53.9/17.8 versus 46.8/27.7. ViT performs better for raw attention, rollout, and CDAM under token deletion. Its best perturbation probe, CDAM at 54.8/17.3, also exceeds GViT Grad-CAM on both measures. ViT rollout has higher segmentation mIoU and mAP than GViT Grad-CAM, while GViT Grad-CAM has higher pixel accuracy. The covariance prior reaches 38.65 mIoU without classifier gradients at evaluation, but its geometry comes from supervised guidance during training. These results characterize the learned interface and its probes; they do not establish uniformly better explanations.

\custompara{Geometric foreground diagnostic.}
We use PASCAL VOC as a diagnostic that asks what kind of layout classifier guidance produces: if guidance reshapes the Gaussian set into a saliency-aligned configuration, then a purely geometric statistic of that set should already correlate with foreground extent on an out-of-distribution benchmark.
Following \citet{bar2022visual} and GMAE~\citep{rajasegaran2025gaussianmae},  we evaluate 
foreground segmentation and single-object detection on PASCAL VOC~\citep{everingham2015pascal}. GMAE represents scenes with layered 3D Gaussians and uses the $z$-axis as depth; rendering individual layers and thresholding yields foreground-background masks {\em for free}, and the \emph{Copy} baseline selects a random true mask from the training set~\citep{bar2022visual}. 
\begin{wraptable}{r}{0.46\textwidth}
% \vspace{-0.32in}
\centering
% \footnotesize
\providecommand{\lightfont}{\color{black!60}}
\renewcommand{\arraystretch}{1.1}
% --- CHANGE 1: Reduced column padding ---
\setlength{\tabcolsep}{1pt} 
\caption{\textbf{Geometric foreground diagnostic} on PASCAL VOC validation, following GMAE~\citep{rajasegaran2025gaussianmae} (IoU; higher is better). GViT uses geometry learned through supervised classifier guidance. Its threshold is selected from the VOC training distribution.
}
\label{tab:pascal_fg}
% --- CHANGE 2: Added S columns for number alignment ---
% \begin{tabular}{l S[table-format=2.2] S[table-format=2.2]} 
\begin{tabular}{l rr} 
\toprule
% --- CHANGE 3: Used \shortstack for multi-line headers ---
\textbf{Model} & {\textbf{\shortstack{Foreg. Segm.}}} & {\textbf{\shortstack{Obj. Det.}}} \\
\midrule
BEiT
& 0.79 & 0.32 \\
MAE
& 4.28 & 1.98 \\
MAE-VQGAN
& 5.26 & 3.21 \\
VQGAN
& 9.13 & 4.99 \\
Copy 
& 14.91 & 13.50 \\
GMAE
& 18.16 & 20.26 \\
\rowcolor{blue!5}
\textbf{GViT~(ours)} & \textbf{26.11} & \textbf{27.36} \\
\bottomrule
\end{tabular}
% \vspace{-0.2in}
\end{wraptable}
In contrast, we keep our representation purely 2D and use the determinant of each Gaussian covariance, $\det(\Sigma)$, as the diagnostic statistic. 
We pool covariance determinants from VOC training images to estimate an empirical cumulative distribution $\widehat F_{\mathrm{train}}$. We retain Gaussians for which $\widehat F_{\mathrm{train}}(\det\Sigma)<0.3$ and render them to form the foreground mask. The training determinant histogram guides selection of the cutoff between its two modes; validation IoU does not select it. The training CDF and cutoff remain fixed on VOC validation (Table~\ref{tab:pascal_fg}). The foreground score is sensitive to the cutoff: CDF thresholds 0.2, 0.3, and 0.4 give 21.06\%, 26.11\%, and 22.14\% IoU. Small Gaussians tend to concentrate in class-relevant regions after guidance. This diagnostic measures foreground structure in the resulting geometry. Its supervised training history differs from that of the self-supervised references, and no unguided VOC result is available to isolate the effect of guidance on this metric.

\begin{table}[t]
  \centering
  \footnotesize
  \setlength{\tabcolsep}{2pt}
    % \vspace{-0.1in}
   \caption{\textbf{Design choices.} Mini-ImageNet top-1 accuracy (\%), reported as mean $\pm$ sample SD over three seeds per setting. Each sweep varies only the named setting; all other parameters use the selected defaults in the paper.}
  \label{tbl:ablations}
  % \vspace{-0.2in}
  \begin{subtable}[t]{0.22\textwidth}
    \centering
    \caption{Scale factor}
    \label{tbl:scale-factor}
    \begin{tabular}{lc}
      \toprule
      Value & top-1 \\
      \midrule
      $c = 0.5$ & $73.47\pm0.21$ \\
      $c = 0.75$ & $74.50\pm0.10$ \\
      $c = 1.0$ & $\mathbf{75.30\pm0.10}$ \\
      $c = 2.0$ & $73.77\pm0.06$ \\
      \bottomrule
    \end{tabular}
    % \vspace{-.1in}
  \end{subtable}%
  \begin{subtable}[t]{0.24\textwidth}
    \centering
    \caption{Guidance}
    \label{tbl:guidance-coeff}
    \begin{tabular}{lc}
      \toprule
      Value & top-1 \\
      \midrule
      $\gamma = 0.025$ & $72.50\pm1.01$ \\
      $\gamma = 0.05$ & $72.87\pm1.54$ \\
      $\gamma = 0.075$ & $73.90\pm0.36$ \\
      $\gamma = 0.1$ & $\mathbf{74.93\pm0.46}$ \\
      \bottomrule
    \end{tabular}
    % \vspace{-.1in}
  \end{subtable}
  \begin{subtable}[t]{0.25\textwidth}
    \centering
    \caption{Reconstruction Loss}
    \label{tbl:loss-function}
    \begin{tabular}{lc}
      \toprule
       Loss     & top-1 \\
      \midrule
      $L_{\text{mse}}$ & $65.30\pm0.36$ \\
      $L_{\text{bce}}$ & $71.57\pm0.42$ \\
      $L_{\text{mse}}+L_{\text{dssim}}$ & $69.47\pm0.35$ \\
      $L_{\text{bce}}+L_{\text{dssim}}$ & $\mathbf{72.47\pm0.15}$ \\
      \bottomrule
    \end{tabular}
    % \vspace{-.1in}
  \end{subtable}%
  \begin{subtable}[t]{0.25\textwidth}
    \centering
    \caption{Gaussian Fitting}
    \label{tbl:method}
    \begin{tabular}{lc}
      \toprule
      Method            & top-1 \\
      \midrule
      Offline & $70.07\pm0.51$ \\
      Learned queries & $71.43\pm0.55$ \\
      Denoising & $74.33\pm0.57$ \\
      Guidance & $\mathbf{75.50\pm0.52}$ \\
      \bottomrule
    \end{tabular}
    % \vspace{-.1in}
  \end{subtable}
  % \vspace{-.1in}
\end{table}

\subsection{Ablations}
\label{sec:ablations}
\vspace{-0.1in}

\custompara{Learning from offline Gaussians.}
A direct baseline fits Gaussians per image with reconstruction-only SGD and trains a classifier on the frozen parameters (Fig.~\ref{fig:bird_splat_comparison}b). The layout is near-uniform, and fitting time and storage grow with dataset size. Feed-forward prediction removes this preprocessing step. Classifier guidance then shapes the predicted layout toward class-relevant regions (Fig.~\ref{fig:bird_splat_comparison}d). In the fitting sweep (Table~\ref{tbl:ablations}d), mean accuracy rises from 70.07\% for offline fitting to 71.43\% for learned queries, 74.33\% for denoising, and 75.50\% with guidance. These results support the combined prediction and guidance design.

\begin{figure}[t!]
  \centering
  \begin{subfigure}[t]{0.49\columnwidth}
    \centering
    \includegraphics[width=\linewidth]{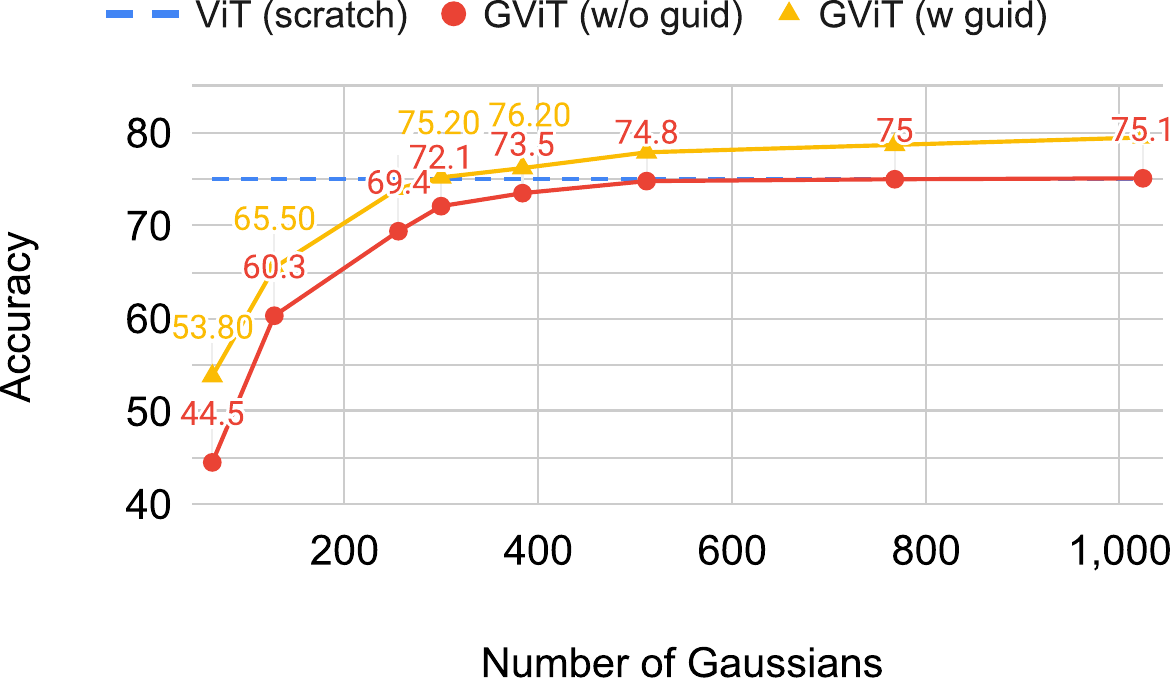}
    \caption{\textbf{Token budget (Mini-ImageNet).} Accuracy increases with $K$ and saturates; guidance peaks near $K=768$.}
    \label{fig:n_gaussians}
  \end{subfigure}\hfill
  \begin{subfigure}[t]{0.49\columnwidth}
    \centering
    \includegraphics[width=\linewidth]{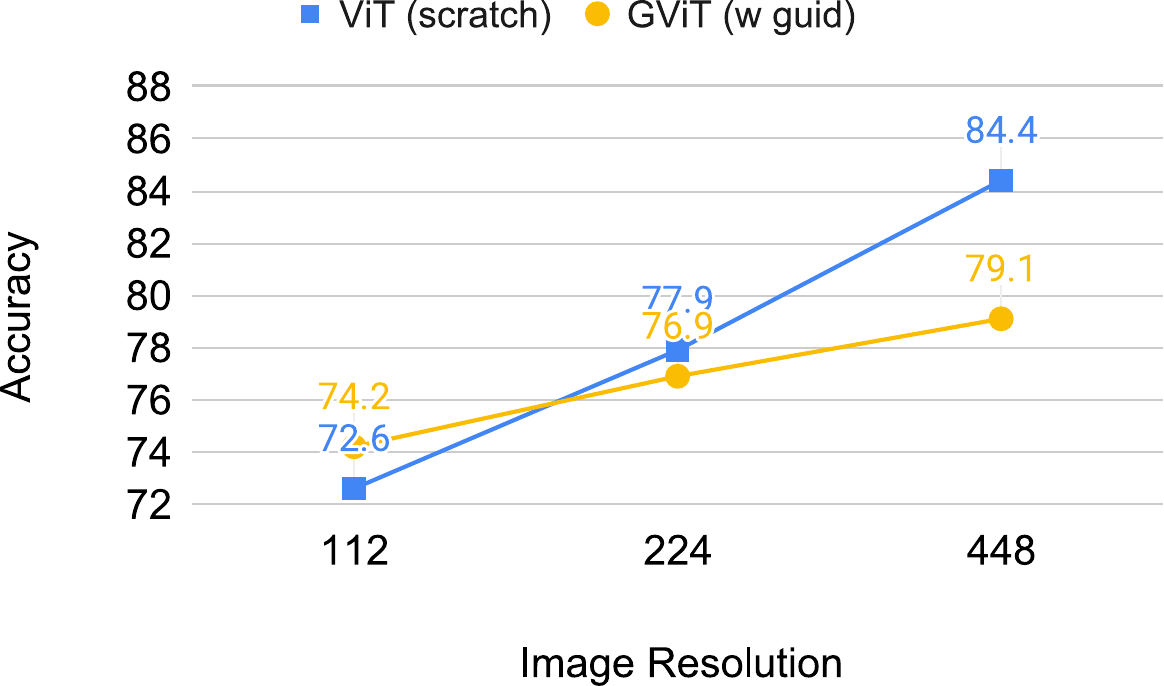}
    \caption{\textbf{Resolution sweep (ImageNet-1K).} Top-1 accuracy across input resolutions.}
    \label{fig:acc_flops_sub}
  \end{subfigure}
% \vspace{-0.1in}
  \caption{\textbf{Token budget and resolution.} (a) Sensitivity to the number of Gaussians. (b) Accuracy across input resolutions for the resolution sweep. The matched-recipe comparisons and inference costs are reported in the text.}
  \label{fig:gaussians_budget_and_efficiency}
  \vspace{-0.2in}
\end{figure}

\custompara{Number of Gaussians.}
We train six variants with token budgets from $K{=}64$ to $K{=}1024$ and report Mini-Imagenet top-1 in Fig.~\ref{fig:n_gaussians}. Without guidance (red), accuracy rises from $44.5\%$ ($K{=}64$) to $75.1\%$ ($K{=}1024$), with diminishing returns beyond roughly $K{=}512$. Guidance (yellow) shifts the curve upward across all budgets: performance exceeds the from-scratch ViT baseline already at $K{=}256$ and peaks at $78.7\%$ around $K{=}768$, then saturates. This indicates that additional Gaussians are most useful when their placement is class-aligned. For ImageNet, we use $K{=}512$ at $224{\times}224$ due to rendering cost.

\custompara{Resolution and inference cost.}
Figure~\ref{fig:acc_flops_sub} shows accuracy across input resolutions. For a matched-resolution comparison at $448\times448$, ViT-B/16 trained with our recipe reaches 84.4\% at 157 GMAC, while GViT-B reaches 79.1\% at 62.8 GMAC. At $224\times224$, the matched-recipe ViT-B/16 reaches 79.8\% at 17.6 GMAC, compared with 76.9\% at 39.2 GMAC for GViT-B. ViT-B/16 at 384 pixels reaches 81.6\% at 55.4 GMAC, exceeding GViT-B at 448 by 2.5 points while using 7.4 fewer GMAC. These operating points compare resolutions and complete systems, without establishing an iso-compute advantage for GViT. The fixed $K=512$ keeps the classifier sequence constant, while the encoder patch sequence still grows with resolution.

\custompara{Gaussian Scale.}
We bound scales with $\hat{s}=c\cdot\mathrm{sigmoid}(s)$. In Table~\ref{tbl:ablations}a, $c=1.0$ has the highest mean accuracy, 75.30\%, compared with 73.47\% at $c=0.5$ and 73.77\% at $c=2.0$. We use $c=1.0$ among these tested settings.

\custompara{Guidance coefficient}
During joint training, $\gamma$ controls the classifier-gradient contribution. Table~\ref{tbl:ablations}b gives the highest mean, $74.93\pm0.46\%$, at $\gamma=0.1$. Extending the sweep to 0.125, 0.15, and 0.2 gives NaNs in 1/3, 1/3, and 2/3 attempts, respectively. Mean accuracy over finite runs is 73.00\%, 71.05\%, and 69.40\%, with only one finite run at 0.2. Failed runs remain part of the stability assessment. Thus, 0.1 is the best tested setting in this sweep, rather than a general optimum.

To separate the guidance signal from extra optimizer updates, we hold the additional encoder-update schedule fixed and set $\gamma=0$. This update-count control reaches $72.03\pm0.12\%$, compared with $74.93\pm0.46\%$ at $\gamma=0.1$, over three seeds each. The 2.90-point mean difference supports a contribution from the classifier-gradient term. This control matches update count, rather than measured training compute, because the zero-guidance update can omit classifier-gradient computation. Figure~\ref{fig:n_gaussians} evaluates $K$ separately. We do not independently sweep the loss weight $\lambda_{\text{perc}}$ or encoder width $d_{\text{enc}}$.

% \begin{figure}[t!]
%   \centering
%   \includegraphics[width=0.68\columnwidth]{figures/n_gaussians.pdf}
%   % \vspace{-0.1in}
%   \caption{\textbf{Number of Gaussians:} Mini-Imagenet classification performance with 64, 128, 256, 512, 768, and 1024 Gaussians per image. 
%   % We evaluate these models on full fine‑tuning. 
%   Performance increases monotonically with more Gaussians.}
%   \label{fig:n_gaussians}
%   \vspace{-0.1in}
% \end{figure}

% Requires: \usepackage{subcaption}

\custompara{Loss function}
We compare MSE, BCE, and their combinations with D-SSIM (Table~\ref{tbl:ablations}c). Mean accuracy is 65.30\% for MSE and 71.57\% for BCE. D-SSIM increases these values to 69.47\% and 72.47\%, respectively. We use BCE with D-SSIM. This comparison measures downstream classification accuracy under each training loss; it does not measure the train-to-validation gap of the losses themselves.

% \custompara{Shortcut diagnostics.}
% Because the classifier reads Gaussian parameters directly, we also test whether the gains could come from non-geometric side channels or stronger texture shortcuts. Cross-training, rasterization controls, and cue-conflict evaluation show no evidence that guidance smuggles in hidden information or increases texture reliance; details are in Appendix~\ref{sec:leakage}.

\section{Conclusion}
\label{sec:conclusions_limits}
\vspace{-0.1in}

\MODEL shows that the same explicit Gaussian parameters can reconstruct an image and serve as classifier tokens. Classifier guidance shapes this representation under a reconstruction constraint. GViT-B reaches 76.9\% on ImageNet-1K at 39.2 GMAC, compared with 79.8\% at 17.6 GMAC for ViT-B/16 trained with our recipe. The explanation probes and covariance diagnostic characterize the spatial structure learned through this joint design. They support inspection of Gaussian geometry while showing that probe performance depends on the method and metric. Potential extensions include pixel or convolutional conditioning, image-dependent primitive budgets, and dense prediction from geometric parameters.

\bibliographystyle{plainnat}
\bibliography{sections/11_references}

\clearpage

\appendix
\counterwithin{figure}{section}
\counterwithin{table}{section}
\renewcommand\thefigure{\thesection.\arabic{figure}}
\renewcommand\thetable{\thesection.\arabic{table}}

\section{Classifier guidance in Gaussian parameter space}\label{app:guidance_intrinsic}\label{sec:linear_guidance_explanation}
\vspace{-0.1in}

This section gives local-linear intuition for guidance. The argument applies to differentiable continuous representations in general. In GViT, the parameters also have explicit geometric and appearance meanings, so their changes can be inspected in image space. Let $\mathbf G\in\mathbb R^{9K}$ contain the Gaussian parameters. Near a reference point $\mathbf G_0$, approximate the classifier logits by
\[
\mathbf z(\mathbf G)\approx\mathbf z(\mathbf G_0)+\mathbf J(\mathbf G-\mathbf G_0),
\qquad
\mathbf p=\operatorname{softmax}(\mathbf z(\mathbf G)).
\]
For cross-entropy and the true-class indicator vector $\mathbf e_y$, the local gradient is $\nabla_{\mathbf G}\mathcal L_{\mathrm{cls}}\approx\mathbf J^\top(\mathbf p-\mathbf e_y)$.

\custompara{Constructive changes in parameter space.}
Let $\mathbf w$ be the row of $\mathbf J$ for the true-class logit. Under a parameter budget $\|\Delta\mathbf G\|_\infty\le\varepsilon$,
\[
\max_{\|\Delta\mathbf G\|_\infty\le\varepsilon}\mathbf w^\top\Delta\mathbf G
=\varepsilon\|\mathbf w\|_1,
\qquad
\Delta\mathbf G^\star=\varepsilon\operatorname{sign}(\mathbf w).
\]
This is the local construction used in adversarial-example intuition~\citep{goodfellow2014explaining}, with the direction chosen to increase the logit. It does not establish a Gaussian-specific advantage or guarantee lower multiclass loss. Patch embeddings also admit classifier gradients. The distinction here is that Gaussian center, scale, and rotation updates directly change spatial support.

\custompara{Reconstruction and recognition.}
Write $\mathbf g_{\mathrm{rec}}=\nabla_{\mathbf G}\mathcal L_{\mathrm{rec}}$ and $\mathbf g_{\mathrm{cls}}=\nabla_{\mathbf G}\mathcal L_{\mathrm{cls}}$. A direct parameter-space step gives the first-order change
\begin{equation}
\label{eq:constructive_regularizer}
\mathcal L_{\mathrm{cls}}\!\left(\mathbf G-\eta(\mathbf g_{\mathrm{rec}}+\gamma\mathbf g_{\mathrm{cls}})\right)
-\mathcal L_{\mathrm{cls}}(\mathbf G)
\approx-\eta\!\left(\mathbf g_{\mathrm{cls}}^\top\mathbf g_{\mathrm{rec}}+\gamma\|\mathbf g_{\mathrm{cls}}\|_2^2\right).
\end{equation}
For positive $\eta$, classification loss decreases locally when the bracketed term is positive. Reconstruction and classification gradients can oppose each other, so reconstruction regularization alone does not guarantee this condition. GViT applies the combined gradient to encoder weights through the predicted parameters (Eq.~\ref{eq:guidance_update}); a direct update of $\mathbf G$ is only an intuition for that procedure. The finite stable guidance range is an empirical result (Sec.~\ref{sec:ablations}). On a random subset of 5,000 ImageNet validation images, guidance changes PSNR/SSIM from 21.9/0.70 to 20.8/0.63, while ImageNet top-1 rises from 73.6\% to 76.9\%. These measurements show a reconstruction--recognition trade-off in the learned layouts.

\section{Limitations}
\label{sec:limitations}

% \MODEL has the following limitations. 
At $224\times224$, GViT-B uses 108M parameters and 39.2 GMAC, compared with about 86M and 17.6 GMAC for ViT-B/16, and has lower ImageNet accuracy. The encoder still consumes image patches, and a fixed $K=512$ controls only the classifier sequence length. The encoder sequence grows with resolution, so the fixed primitive budget alone does not establish better scaling of full-model compute. Table~\ref{tab:modalityin1k} compares systems with different capacities and training settings; a controlled comparison with another structured tokenizer remains open. Training uses three phases and additional guidance updates. Our ablations report three seeds per setting, but they do not independently sweep the loss weight or encoder width, and we do not report seed variation for the headline ImageNet comparison. Matched training-compute measurements and train/validation loss curves are also outside this evaluation.

Our evaluation covers classification, transfer, and diagnostic localization. Gaussian geometry supports direct projection of relevance, but segmentation overlap and token deletion remain imperfect proxies for explanation faithfulness. The deletion protocol retains Gaussian centers and intervenes after encoding; it does not match removed image area or measure pixel deletion followed by re-encoding. The VOC diagnostic uses geometry learned with supervised guidance and a threshold selected on VOC training data. It does not isolate an independent unsupervised foreground capability. Dense prediction, self-supervised learning, video, and adaptive primitive budgets remain possible extensions.

\section{Implementation Details}\label{supp:impl}
\vspace{-0.1in}

\custompara{Overall architecture.}

Here, we use a notation more amenable to computation than the one used in the main paper.
\MODEL\ consists of three modules that are trained jointly: 
(i) a Denoising Gaussian Encoder that maps an image to a compact set of 2D Gaussian parameters, 
(ii) a differentiable Gaussian renderer that turns these primitives back into an RGB image, and 
(iii) a ViT-style classifier that operates directly on the Gaussian parameters. 
Given an input image $I \in \mathbb{R}^{3\times H\times W}$ we predict $K$ Gaussians.
\[
g_j = \{p_j, s_j, \alpha_j, r_j, o_j\} \in \mathbb{R}^9,
\]
with $ j=1,\dots,K$ and where $p_j \in [-1,1]^2$ are 2D centers, $s_j \in \mathbb{R}^2$ are scale logits, $\alpha_j$ is an in-plane rotation angle, $r_j \in [0,1]^3$ are RGB colors and $o_j \in [0,1]$ is opacity.
Unless otherwise stated we use $K{=}512$ Gaussians at $224{\times}224$ resolution.
All Gaussians are constrained to lie on the $z{=}0$ plane and only rotate in the $x$–$y$ plane (see Algorithm~\ref{alg:method}), so the representation is purely 2D even when using a 3D renderer. Concretely, we implement 2D Gaussians via a constrained 3D splatting backend (\texttt{gsplat}~\citep{ye2025gsplat}): the 9D parameterization $(p,s,\alpha,r,o)$ of the main paper is mapped to a 14D internal vector $[\,\mu\!\in\!\mathbb{R}^3,\,q\!\in\!\mathbb{R}^4,\,\sigma\!\in\!\mathbb{R}^3,\,o\!\in\!\mathbb{R},\,r\!\in\!\mathbb{R}^3\,]$ with $\mu_z{=}0$ and the $x,y$ components of the quaternion $q$ zeroed, so the renderer is 3D in form but 2D in degrees of freedom.

\custompara{Denoising Gaussian Encoder.}

The encoder is ViT-S-sized, with about 22M parameters, and operates on image patches and random Gaussian inputs.
We split the image into non-overlapping patches, project each patch to a $d_{\text{enc}}$-dimensional embedding.
We then sample $K$ random Gaussians $g_0 \in \mathbb{R}^{K\times 9}$ once per iteration and linearly project them into the same token space.
The final encoder input sequence is
\[
[\texttt{patch}_1, \dots, \texttt{patch}_N, \texttt{gauss}_1, \dots, \texttt{gauss}_K] + E_{\text{pos}},
\]
where $E_{\text{pos}}$ is a learned positional embedding covering both patch and Gaussian tokens.
The transformer stack (8 blocks in our ImageNet experiments) outputs a sequence of the same length; we discard the patch tokens and keep only the $K$ Gaussian latents.
A two-layer MLP head predicts residuals $\Delta g$, which are added to the random initialization $g_0$:
\[
\hat{g} = g_0 + \Delta g.
\]
This “one-step denoising’’ view lets the encoder reshape an initial random Gaussian layout into a task-adapted mid-level representation.

\custompara{Gaussian renderer.}
The renderer decodes $\hat{g}$ into differentiable Gaussian parameters and splats them into an image.
In the default 2D setting, each 9D vector is split as
\[
(p, s, \alpha, r, o) = \text{split}(\hat{g}, [2,2,1,3,1]).
\]
We map $p$ to normalized image coordinates via $\tanh$, bound the scales with $s_{\text{pos}} = c\cdot\sigma(s)$ where $c$ is a global scale cap (Table~\ref{tab:impl_gvit_pretrain}) and $\sigma$ is the logistic sigmoid, and squish colors and opacities through sigmoids to $[0,1]$.
We use the 2D Gaussian splatting backend from \texttt{gsplat}~\citep{ye2025gsplat}, which directly rasterizes these 2D Gaussians on a fixed canvas.
We constrain the means to the $z{=}0$ plane and zero out the $x$ and $y$ components of the quaternion, so that only in-plane rotations are allowed.
The renderer returns a reconstructed image $\hat{I}$ of the same spatial size as the input.

\custompara{Gaussian classifier.}
The classifier treats each Gaussian as a token and applies a ViT-B style backbone.
For 2D Gaussians we first construct a feature vector from the raw parameters:
centers $\boldsymbol\mu$, $\log$-scales, $\sin$/$\cos$ of the angle $\alpha$, colors, and opacity.
These are concatenated, linearly projected to a classifier width $d_{\text{cls}}$, and augmented with a learnable class token and positional embeddings over Gaussians.
\begin{table}[t!]
% \scriptsize
\centering
\caption{\textbf{Implementation details.} \textit{Left:} ImageNet‐1k pre-training settings. \textit{Right:} fine-tuning hyper-parameters for downstream transfer.}
\label{tab:impl_gvit_all}
\begin{minipage}[t]{0.48\linewidth}
\centering
\subcaption{\textbf{ImageNet-1K Pre-training}}
\label{tab:impl_gvit_pretrain}
\begin{tabular}{lc}
\toprule
config & value \\ \midrule
optimizer & AdamW  \\
base learning rate & 1e-4 \\
weight decay & 0.05 \\
optimizer momentum & $\beta_{1,2}{=}0.9,0.95$  \\
batch size & 4096 \\
lr schedule & cosine decay  \\
warm-up epochs & 10 \\
epochs & 400 \\
augmentation & hflip, crop [0.5, 1] \\
$\lambda_{\text{perc}}$ & 0.1 \\
perc-loss schedule & $0{\rightarrow}0.1$ (from epoch 150) \\
$\lambda_{\text{cls}}$ & 1 \\
\bottomrule
\end{tabular}
\end{minipage}%
\hfill
\begin{minipage}[t]{0.45\linewidth}
\centering
\subcaption{\textbf{Transfer to Other Datasets}}
\label{tab:impl_gvit_finetune}
\begin{tabular}{lcc}
\toprule
config & ViT/S & ViT/B \\
\midrule
optimizer & \multicolumn{2}{c}{AdamW} \\
optimizer momentum & \multicolumn{2}{c}{$\beta_1, \beta_2{=}0.9, 0.999$} \\
base learning rate  \\
\quad {C10, C100} & 1e-3 & 3e-3 \\
\quad {CUB} & \multicolumn{2}{c}{4.8e-3} \\
\quad {SUN} & 3.4e-3 & 4.8e-3 \\
\quad {Pets, Flowers} & \multicolumn{2}{c}{1e-3} \\
\quad {DTD} & \multicolumn{2}{c}{3e-3} \\
weight decay & \multicolumn{2}{c}{0.05} \\
learning rate schedule &  \multicolumn{2}{c}{cosine decay} \\
warm-up epochs & \multicolumn{2}{c}{5} \\
layer-wise lr decay & 0.65 & 0.75 \\
batch size & 1024 & 768 \\
training epochs \\
\quad {C10, C100} & 60 & 50 \\
\quad {CUB, SUN} & 100 & 80 \\
\quad {Pets, Flowers} & 100 & 50 \\
\quad {DTD} & \multicolumn{2}{c}{50} \\
augmentation & \multicolumn{2}{c}{hflip, crop [0.5, 1]} \\
\bottomrule
\end{tabular}
\end{minipage}
\end{table}
% \begin{table}[t!]
% \small
% \centering
% \caption{\textbf{ImageNet-1K Pre-training.} Implementation details and settings.}
% \label{tab:impl_gvit_pretrain}
% \begin{tabular}{lc}
% \toprule
% config & value \\ \midrule
% optimizer & AdamW \citep{loshchilov2017decoupled} \\
% base learning rate & 1e-4 \\
% weight decay & 0.05 \\
% optimizer momentum & $\beta_{1,2}{=}0.9,0.95$ \citep{chen2020generative} \\
% batch size & 4096 \\
% lr schedule & cosine decay \citep{loshchilov2016sgdr} \\
% warm-up epochs & 10 \citep{goyal2017accurate} \\
% epochs & 400 \\
% augmentation & hflip, crop [0.5, 1] \\
% $\lambda_{\text{perc}}$ & 0.1 \\
% perc-loss schedule & $0{\rightarrow}0.1$ (from epoch 150) \\
% $\lambda_{\text{cls}}$ & 1 \\
% \bottomrule
% \end{tabular}
% \end{table}
We use a 12-layer Transformer with multi-head self-attention and a standard MLP block per layer, followed by a final LayerNorm and a linear classification head on the class token.
To better exploit the explicit spatial structure of the Gaussians, we equip the attention layers with a simple 2D rotary positional embedding (RoPE) applied to query/key vectors, using the normalized $(x,y)$ centers as positions.

\custompara{Training objective and gradient guidance.}
Training uses three losses: a pixel-wise reconstruction loss $\mathcal{L}_{\text{pix}}(\hat{I}, I)$, a perceptual reconstruction loss $\mathcal{L}_{\text{perc}}(\hat{I}, I)$ based on DSSIM, and a standard cross-entropy classification loss $\mathcal{L}_{\text{cls}}$ on the classifier logits.
We combine them as
\[
\mathcal{L}_{\text{rec}} = \mathcal{L}_{\text{pix}} + \lambda_{\text{perc}}\,\mathcal{L}_{\text{perc}}, 
\qquad
\mathcal{L} = \mathcal{L}_{\text{rec}} + \lambda_{\text{cls}}\,\mathcal{L}_{\text{cls}}.
\]
By default we use a pixel-wise BCE loss for $\mathcal{L}_{\text{pix}}$ and set $\lambda_{\text{cls}}{=}1$ and $\lambda_{\text{perc}}{=}0.1$.
Algorithm~\ref{alg:method} summarizes the full training loop, including the \emph{guidance} step in which classifier gradients are reused to relocate Gaussians in a constructive FGSM-style fashion.
Concretely, we accumulate encoder gradients from the reconstruction objective ($\nabla_{\phi} \mathcal{L}_{\text{rec}}$) and from the classification objective ($\nabla_{\phi} \mathcal{L}_{\text{cls}}$), and combine them into a single \emph{gradient passed to a minimization optimizer}
\[
\tilde{\nabla}_\phi = \nabla_{\phi}\mathcal{L}_{\text{rec}} + \gamma\,\nabla_{\phi}\mathcal{L}_{\text{cls}},
\]
where $\phi$ denotes encoder parameters and $\gamma$ is the guidance coefficient (Eq.~(1) in the main paper).
The optimizer then subtracts $\eta\,\tilde{\nabla}_\phi$ from $\phi$, performing simultaneous gradient descent on $\mathcal{L}_{\text{rec}}$ and $\mathcal{L}_{\text{cls}}$; this keeps reconstructions faithful while nudging Gaussians toward features that increase the true-class logit.

We implement the full architecture in Pytorch.
The encoder and classifier are optimized with separate AdamW~\citep{loshchilov2017decoupled} optimizers but share the same hyperparameters.
We use mixed-precision training (bfloat16) and gradient clipping at 0.5, and rely on \texttt{gsplat}~\citep{ye2025gsplat} for 2D Gaussian splatting~\citep{huang20242d}.

% \begin{table}[t!]
% \small
% \centering
% \caption{\textbf{Transfer to Other Datasets.} Fine-tuning hyper-parameters for downstream transfer.}
% \label{tab:impl_gvit_finetune}
% \begin{tabular}{lcc}
% \toprule
% config & \MODEL-S & \MODEL-B \\
% \midrule
% optimizer & \multicolumn{2}{c}{AdamW} \\
% optimizer momentum & \multicolumn{2}{c}{$\beta_1, \beta_2{=}0.9, 0.999$} \\
% base learning rate & & \\
% \quad {C10, C100} & 1e-3 & 3e-3 \\
% \quad {CUB} & \multicolumn{2}{c}{4.8e-3} \\
% \quad {SUN} & 3.4e-3 & 4.8e-3 \\
% \quad {Pets, Flowers} & \multicolumn{2}{c}{1e-3} \\
% \quad {DTD} & \multicolumn{2}{c}{3e-3} \\
% weight decay & \multicolumn{2}{c}{0.05} \\
% learning rate schedule & \multicolumn{2}{c}{cosine decay} \\
% warm-up epochs & \multicolumn{2}{c}{5} \\
% layer-wise lr decay \citep{clark2020electra,bao2021beit} & 0.65 & 0.75 \\
% batch size & 1024 & 768 \\
% training epochs & & \\
% \quad {C10, C100} & 60 & 50 \\
% \quad {CUB, SUN} & 100 & 80 \\
% \quad {Pets, Flowers} & 100 & 50 \\
% \quad {DTD} & \multicolumn{2}{c}{50} \\
% augmentation & \multicolumn{2}{c}{hflip, crop [0.5, 1]} \\
% \bottomrule
% \end{tabular}
% \end{table}

\custompara{ImageNet-1K pre-training.}
The default ImageNet-1K settings are summarized in Table~\ref{tab:impl_gvit_pretrain}.
We use AdamW with base learning rate $10^{-4}$ and weight decay $0.05$, cosine learning-rate decay, and a batch size of 4096.
We follow the common linear learning-rate scaling rule $\text{lr} = \text{base\_lr} \times \text{batchsize} / 256$~\citep{goyal2017accurate}.
Data augmentation is intentionally simple: random resized crops with scale in $[0.5, 1]$ and random horizontal flipping, with color jitter, and Mixup.
We train for 400 epochs with a three-phase schedule:
(i) a reconstruction warm-up, where only the encoder and renderer are optimized using $\mathcal{L}_{\text{rec}}$ (no classification or guidance);
(ii) a classifier warm-up phase in which the encoder is frozen and the ViT classifier is trained on the (currently) learned Gaussians with $\mathcal{L}_{\text{cls}}$;
and (iii) a joint phase in which both encoder and classifier are updated with the full loss and gradient guidance enabled.
On ImageNet-1K we allocate 150, 100, and 150 epochs to these three phases, respectively, use $\lambda_{\text{perc}}{=}0.1$ during the joint phase, and set the guidance coefficient to $\gamma{=}0.1$ with one guidance step after every 10 normal updates.
The \emph{no-guidance} variant uses the same schedule but omits the guidance step in phase (iii).

\custompara{Transfer to other datasets.}
For downstream benchmarks, we start from the ImageNet-pretrained \MODEL\ and \emph{freeze the Gaussian encoder}, fine-tuning only the ViT classifier on each target dataset with a purely discriminative objective.
The number of Gaussians, the Gaussian parameterization, and the renderer are kept fixed.
Table~\ref{tab:impl_gvit_finetune} lists dataset-specific learning rates, batch sizes, and training epochs for the GViT-S and GViT-B variants.
We follow common ViT practice and apply layer-wise learning-rate decay~\citep{clark2020electra,bao2021beit} with decay factors $0.65$ (GViT/S) and $0.75$ (GViT/B), use cosine learning-rate decay with a short warm-up, and keep the same crop and flip augmentation (random resized crop and horizontal flip).
All reported transfer results are obtained from a single linear classification head on top of the \MODEL\ class token.

\section{Extra results.}\label{app:extra_results}
\vspace{-0.1in}

\begin{table}[t!]
\centering
\footnotesize
\setlength{\tabcolsep}{3pt}
\renewcommand{\arraystretch}{1.12}
\caption{\textbf{Classifier input interfaces and complete-system comparisons.} ImageNet-1K top-1 accuracy (\%). Methods differ in capacity, resolution, token budget, and training. Token counts exclude CLS. GMAC counts one multiply-accumulate once; costs cover the stated model operations, not a uniform end-to-end image pipeline. Dashes indicate unavailable counts. Panel (b) separates tokenizer parameters from the remaining model.}
\label{tab:modalityin1k}
\textbf{(a) Representation, capacity, and inference cost}\par\smallskip
\begin{tabular}{@{}>{\raggedright\arraybackslash}p{3.4cm}>{\raggedright\arraybackslash}p{3.0cm}r >{\raggedright\arraybackslash}p{3.6cm}r r r@{}}
\toprule
\textbf{Method} & \textbf{Representation} & \textbf{Res.} & \textbf{Token sequence} & \shortstack{\textbf{Params}\\\textbf{(M)}} & \textbf{GMAC} & \textbf{Top-1} \\
\midrule
DeiT-Ti/16~\citep{pmlr-v139-touvron21a} & RGB patches & 224 & 196 & 5.72 & 1.25 & 72.2 \\
DeiT-S/16 & RGB patches & 224 & 196 & 22.05 & 4.60 & 79.8 \\
DeiT-B/16 & RGB patches & 224 & 196 & 86.57 & 17.56 & 81.8 \\
ByteFormer-Ti~\citep{horton2023bytes} & JPEG bytes; windowed attention & 224 & Variable; $\approx6{,}069$ initially, then halving & 16.22 & $\approx11.57$ & 64.9 \\
JPEG-Ti~\citep{park2023rgb} & Grouped JPEG DCT & 224 & 196 & 5.64 & 1.23 & 75.1 \\
JPEG-S & Grouped JPEG DCT & 224 & 196 & 21.98 & 4.55 & 76.5 \\
SuiT-Ti~\citep{lew2024superpixel} & Superpixels & 224 & $\le196$; mean $\approx192$ & 5.5 & 1.44 & 75.7 \\
SuiT-S & Superpixels & 224 & $\le196$; mean $\approx192$ & 21.7 & 5.20 & 80.9 \\
SuiT-B & Superpixels & 224 & $\le196$; mean $\approx192$ & 86.0 & 19.7 & 82.3 \\
DART + DeiT-Ti~\citep{yin2025dart} & Adaptive resampled patches & 448 & 196 & 7.0 & -- & 73.8 \\
DPT-Ti~\citep{chen2021dpt} & Deformable patches; hierarchical PVT & 224 & $3{,}136\!\to\!784\!\to\!196\!\to\!49$ & 15.15 & $\approx2.14$ & 77.4 \\
Perceiver~\citep{jaegle2021perceiver} & Pixels; learned positions & 224 & $50{,}176$ inputs $\to512$ latents & 55.9 & 202 & 67.6 \\
Perceiver IO~\citep{jaegle2022perceiver} & Pixels; learned positions & 224 & $50{,}176$ inputs $\to512$ latents & 62.3 & 203.5 & 72.7 \\
Pixel-ViT-B/1~\citep{nguyen2024image} & Individual RGB pixels & 28 & 784 & $\approx86$ & $\approx78$ & 76.1 \\
\rowcolor{blue!5}
\MODEL-B & Patch-conditioned Gaussians & 224 & Encoder: $196+512$; classifier: 512 & $22+86$ & 39.2 & 76.9 \\
\bottomrule
\end{tabular}
\par\medskip
\textbf{(b) Parameter split and training behind the quoted result}\par\smallskip
\begin{tabular}{@{}>{\raggedright\arraybackslash}p{3.4cm}>{\raggedright\arraybackslash}p{6.6cm}>{\raggedright\arraybackslash}p{5.5cm}@{}}
\toprule
\textbf{Method} & \textbf{Tokenizer + remaining parameters (M)} & \textbf{Training} \\
\midrule
DeiT-Ti / S / B & Ti: $0.148+5.570$; S: $0.295+21.755$; B: $0.591+85.977$ & 300 epochs; nondistilled \\
ByteFormer-Ti & $10.239+5.976$ & 300 epochs; JPEG k16 configuration \\
JPEG-Ti / S & Ti: $0.074+5.569$; S: $0.148+21.827$ & 300 / 90 epochs \\
SuiT-Ti / S / B & Ti: $0.017+5.532$; S: $0.052+21.680$; B: $0.177+85.826$ & 300 epochs; from scratch \\
DART + DeiT-Ti & Exact split unavailable; 7M total & 300 epochs; frozen ImageNet-pretrained scorer \\
DPT-Ti & $2.782+12.371$ & 300 epochs \\
Perceiver / IO & $\approx12.9+43.0$ / $\approx12.9+49.4$ & 110 epochs \\
Pixel-ViT-B/1 & Exact split unavailable & 300 epochs \\
\rowcolor{blue!5} \MODEL-B & $22+86$ & 400 epochs across three phases \\
\bottomrule
\end{tabular}
\end{table}

\custompara{Comparison to alternative input representations.}
Table~\ref{tab:modalityin1k} reports complete systems, including fixed-patch references. Matching the classifier backbone alone does not match encoder cost, sequence length, input resolution, or training. For example, SuiT-B reaches 82.3\% with about 86M parameters and 19.7 GMAC, while GViT-B reaches 76.9\% with 108M parameters and 39.2 GMAC. These results expose the performance gap. Our question is whether the parameters of a renderable geometric image model can themselves support classification under task-driven guidance. SuiT pools pixel features, and DART and DPT embed adaptively sampled regions. GViT classifies the Gaussian parameters that also reconstruct the image. This distinction motivates the representation; the table does not establish that Gaussians outperform other adaptive tokenizers under controlled conditions.

\custompara{Cost accounting.}
DeiT and JPEG parameter splits and GMAC, DPT dense operations, Pixel-ViT compute, and SuiT parameter splits are architecture-based counts. ByteFormer uses the k16 JPEG configuration at a stipulated 48,564-byte input; its derived cost is not a dataset-average measurement, and its tokenizer split includes a 50,000-entry positional table. SuiT's reported GMAC does not establish inclusion of CPU superpixel clustering; its mean region count is not a fixed dense-attention length. DPT's derived count excludes custom sampling. DART's normalized full-system count is unavailable. Perceiver costs convert the source's two-operation convention to GMAC, with approximate parameter splits from rounded totals; the source count and released learned-position configuration differ. GViT costs use the measured encoder-plus-classifier forward pass. These scope differences limit comparisons of end-to-end computational cost.

\custompara{Results for ViT-S and \MODEL-S.}
Table~\ref{tab:main_result_small} reports the smaller classifier setting. Guided GViT-S reaches 72.4\% top-1, compared with 75.9\% for ViT-S/16 trained with our recipe. The full GViT-S system has approximately $22+22=44$M parameters and requires 24.2 GMAC at 224 pixels, compared with about 22M and 4.6 GMAC for ViT-S/16. Guidance improves ImageNet accuracy from 68.3\% to 72.4\% and the seven-dataset transfer mean from 81.3\% to 83.8\%. These results show the same accuracy and cost limitations at the Small classifier scale. Nondistilled DeiT-S/16 provides a stronger 79.8\% literature reference~\citep{pmlr-v139-touvron21a}.

\custompara{Reusing a \MODEL-S encoder for \MODEL-B.}
We also evaluate reusing the \emph{same} Gaussian encoder trained jointly with a \MODEL-S classifier when scaling up to \MODEL-B. In this setting, we keep the encoder fixed, replace only the classifier with the larger GViT-B head, and resume training with a short fine-tuning phase. For stability, we lower the guidance weight from $\gamma{=}0.10$ to $\gamma{=}0.05$ during this stage. This simple shortcut reaches 74.5\% top-1 accuracy on ImageNet-1K, only a modest drop compared to the fully trained \MODEL-B with guidance (76.9\%), while roughly halving the end-to-end training time required for the large model.

\begin{table*}[t]
% \vspace{-0.1in}
  \newcommand{\grayfont}{\color{black!60}}
  \centering
  % \footnotesize
  \renewcommand{\arraystretch}{1.5}
  \setlength{\tabcolsep}{1.5pt}
  \caption{Top-1 classification accuracies (\%) on ImageNet-1K (IN-1K) and seven transfer datasets for Small classifiers. Dataset abbreviations and AVG follow Table~\ref{tab:main_result}: AVG uses only positive values in the transfer columns and excludes IN-1K and missing entries. Literature ViT-S baselines use their source protocols~\citep{dosovitskiy2020image,steiner2022how}.}
  \label{tab:main_result_small}
  \begin{tabular}{l l c c c c c c c c c}
    \toprule
    \multirow{2}{*}{\textbf{Model}} & 
    \multirow{2}{*}{\textbf{Input}} &
    & \multicolumn{8}{c}{\textbf{Fine-grained Benchmarks}} \\
    \cmidrule(lr){4-11}
    & & \textbf{IN-1K} & \textbf{AVG} & \textbf{C10} & \textbf{C100} & \textbf{CUB} & \textbf{SUN} & \textbf{Flwr} & \textbf{Pets} & \textbf{DTD} \\
    \midrule
    \grayfont ViT-S/32~\citep{dosovitskiy2020image} & Patches & \grayfont 69.2 & \grayfont 84.6 & \grayfont - & \grayfont 86.4 & \grayfont 92.7 & \grayfont 72.9 & \grayfont 93.6 & \grayfont 91.2 & \grayfont 70.7 \\
    \grayfont ViT-S/16~\citep{steiner2022how} & Patches & \grayfont 77.5 & \grayfont 87.8 & \grayfont 97.9 & \grayfont 86.9 & \grayfont 93.1 & \grayfont 74.3 & \grayfont 95.7 & \grayfont 93.8 & \grayfont 72.8 \\
    \midrule
    \MODEL-S {(no guid)} & {Gaussians} & 68.3 & 81.3 & 92.3 & 79.4 & 86.3 & 63.9 & 89.3 & 86.4 & 71.4 \\
    \rowcolor{blue!5}
    \MODEL-S {(w/guid)} & {Gaussians} & {72.4} & 83.8 & {95.3} & {80.2} & {88.5} & {65.9} & {94.8} & {89.0} & {72.9} \\
    \bottomrule
  \end{tabular}
  % \vspace{-0.2in}
\end{table*}

\section{Shortcut diagnostics}
\label{sec:leakage}
\vspace{-0.1in}

The main paper argues that classifier guidance produces task-aligned Gaussian layouts. A natural concern is whether the same gains could instead come from non-geometric side channels: imperceptible cues encoded in the predicted parameters, idiosyncrasies of a particular Gaussian set, or texture shortcuts that any ImageNet-trained model can fall into. The three diagnostics in this section ask what content the classifier is actually using, by varying the representation it consumes (cross-training and rasterization) and stress-testing it against an external probe (shape vs.\ texture bias). None of these is a proof of faithfulness; together they characterize the kind of evidence the classifier relies on.

\begin{table}[t]
\centering
% \scriptsize
\renewcommand{\arraystretch}{1.1}
\setlength{\tabcolsep}{6pt}
\caption{\textbf{Shortcut diagnostics.} (a) Comparable GS/IMG rankings across training settings argue against hidden information leakage. (b) GViT is substantially more shape-biased than ResNet-50 and close to patch-based ViTs under the \citet{geirhos2018imagenet} cue-conflict test.}
\label{tab:shortcut_diagnostics}
% \vspace{-0.2in}
\begin{subtable}[t]{0.45\linewidth}
\centering
\caption{Leakage check: Mini-ImageNet accuracy training on Gaussians (GS) vs renderings (IMG).}
\label{tab:leakage_render}
\begin{tabular}{l S[table-format=2.1] S[table-format=2.1]}
  \toprule
  \textbf{Setup} & {\textbf{GS}} & {\textbf{IMG}}\\
  \midrule
  SGD Gaussians        & 70.5 & 68.5\\
  Learned Gaussians    & 67.6 & 65.0\\
  Denoising training   & 71.4 & 70.5\\
  Guidance training    & \textbf{72.1} & \textbf{71.0}\\
  \bottomrule
\end{tabular}
\end{subtable}\hfill
\begin{subtable}[t]{0.45\linewidth}
\centering
\caption{Cue-conflict shape bias~\citep{geirhos2018imagenet} (higher = more shape-biased).}
\label{tab:cue_conflict_shape_bias}
\begin{tabular}{l S[table-format=1.2]}
  \toprule
  \textbf{Model} & {\textbf{Shape bias} $\uparrow$}\\
  \midrule
  ResNet-50 & 0.21\\
  % DeiT-S~\citep{pmlr-v139-touvron21a}    & 0.30\\
  DeiT-B    & 0.39\\
  % ViT-S/16~\citep{dosovitskiy2020image}  & 0.38\\
  ViT-B/16  & 0.50\\
  \midrule
  % \textbf{GViT-S} & \textbf{0.32}\\
  \textbf{GViT-B} & \textbf{0.45}\\
  \bottomrule
\end{tabular}
\end{subtable}
% \vspace{-0.2in}
% \vspace{-0.32in}
\end{table}

\custompara{Experiment 1 -- Cross-training between Gaussian sets.}
This experiment asks whether two procedurally distinct Gaussian sets carry compatible class information, or whether each classifier latches onto encoding-specific artifacts. We generate two frozen collections of Gaussians: \emph{SGD} Gaussians obtained by running an image-reconstruction optimizer on each training image, and \emph{learned} Gaussians produced by our gradient-guided encoder. A ViT-S classifier trained on the SGD Gaussians reaches \textbf{70.5\%} Mini-Imagenet accuracy on its native representation and \textbf{67.4\%} when evaluated on learned Gaussians; conversely, the classifier trained on the learned representation reaches \textbf{71.4\%} on its own Gaussians and \textbf{68.5\%} on SGD Gaussians. The diagonal is strong and the $\sim$3-point off-diagonal drop is symmetric, which is more consistent with both sets encoding similar class-relevant structure than with each classifier exploiting an encoding-specific shortcut.

\custompara{Experiment 2 -- Rasterization control.}
This experiment asks whether the signal the classifier uses survives a lossy round-trip through pixels, a representation that strips any non-rendered side channel encoded in the raw parameters. We compare classifiers trained on Gaussian parameters (GS) against the same classifiers fed renderings of those Gaussians (IMG). Table~\ref{tab:leakage_render} shows that GS$\to$IMG scores fall by a modest $2$--$3$ points without changing the relative ranking of training settings. We read this as no evidence of a \emph{large} non-rendered shortcut: most of the class-relevant content is recoverable from the rendered image, and direct access to parameters offers a small, consistent margin on top.

\custompara{Experiment 3 -- Shape vs.\ texture bias (cue-conflict).}
As an external sanity check, we run the cue-conflict evaluation of \citet{geirhos2018imagenet}, in which each image carries a shape label and a conflicting local texture label, and report \emph{shape bias} (the fraction of predictions matching the shape among those matching either label). Table~\ref{tab:cue_conflict_shape_bias} places GViT-B at $0.45$, between ResNet-50 ($0.21$) and ViT-B/16 ($0.50$) and close to DeiT-B ($0.39$). We use this as reassurance rather than as a headline contribution: it indicates that swapping patch tokens for guided Gaussian primitives does not introduce a stronger texture shortcut than the patch-based ViTs we compare against, and the representation remains meaningfully more shape-driven than CNNs.

Taken together, these diagnostics are consistent with the co-design view of Section~\ref{sec:method}: the diagnostics support task-aligned Gaussian layouts without evidence of a large non-rendered shortcut. They do not prove that information cannot spread across tokens or explain the full guidance gain. The extra-update control in Sec.~\ref{sec:ablations} tests a separate optimization confound.

\section{More qualitative examples.}
\vspace{-0.1in}

\begin{landscape}
\begin{figure}[p]
  \centering
  \includegraphics[width=\linewidth]{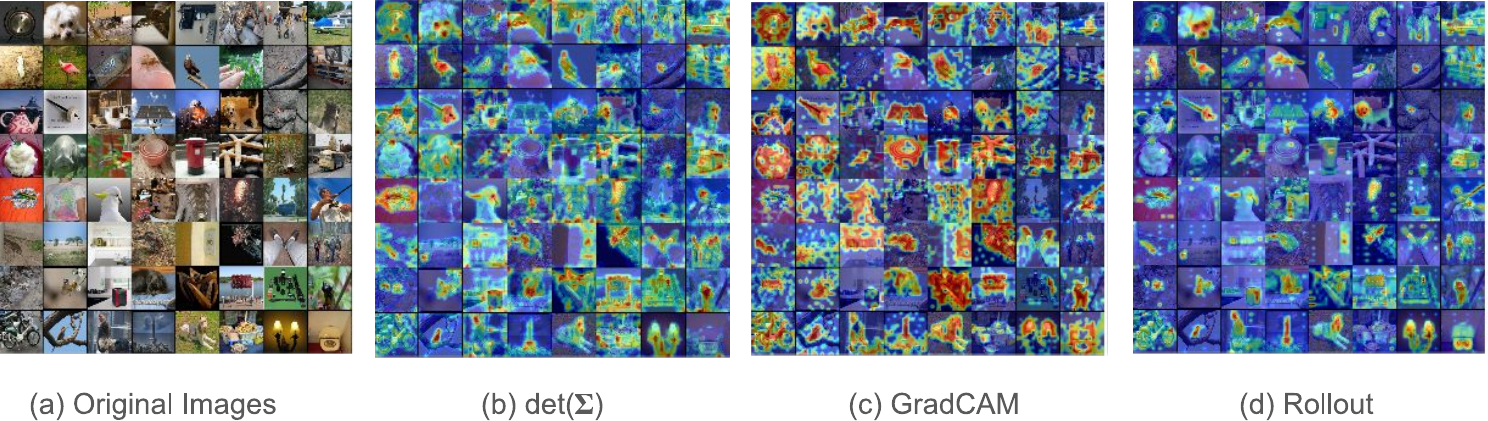}
  \caption{Gradient‐guided Gaussians exhibit a natural interpretability component. (a) Original input images. (b) Determinant of the covariance (\(\det(\Sigma)\)), which clusters Gaussian primitives in class‐salient regions. (c) GradCAM over the learned Gaussians. (d) Rollout of the attentions.}
  \label{fig:interpretability2}
\end{figure}
\end{landscape}

In Fig.~\ref{fig:interpretability2}, we provide additional qualitative examples on ImageNet-1K.
For each validation image, we visualize: (a) the original RGB input; (b) the determinant of the covariance $\det(\Sigma)$ for each Gaussian, rendered as a heatmap over the image; (c) GradCAM over the learned Gaussians; and (d) the rollout of the attention maps.
These examples illustrate how the gradient-guided encoder tends to allocate small, high-precision Gaussians to class-salient structures (such as heads, limbs, or distinctive textures) while using larger Gaussians to model background context.
The GradCAM visualizations typically peak in similar regions, and the reconstructions show that this sparse, mid-level representation is sufficient to preserve the overall layout and colors of the image.
All examples are raw model outputs, without cherry-picking.

\begin{algorithm}[t!]
% \scriptsize
  \caption{\MODEL \ PyTorch pseudocode.}
  \label{alg:method}
    \definecolor{codeblue}{rgb}{0.25,0.5,0.5}
    \definecolor{codekw}{rgb}{0.85, 0.18, 0.50}
    \providecommand{\algofontsize}{7pt}
    \lstset{
      backgroundcolor=\color{white},
      basicstyle=\fontsize{\algofontsize}{\algofontsize}\ttfamily\selectfont,
      columns=fullflexible,
      breaklines=true,
      captionpos=b,
      commentstyle=\fontsize{\algofontsize}{\algofontsize}\color{codeblue},
      keywordstyle=\fontsize{\algofontsize}{\algofontsize}\color{codekw},
    }
\begin{lstlisting}[language=python]
# I[N, C, H, W] - minibatch
# plambda: perceptual coefficient
# clambda: class coefficient
# gamma: guidance coefficient
# guide_every: guidance interval, e.g. 10 steps
# scale_factor: scale factor
# fixed_z_scale: backend-only scale for unused z dimension
# kwargs: focal_length, viewmat, Ks, other gsplat args

def forward_gvit(I):
    g0 = random_gaussian(batch_size=I.shape[0], K=K, device=I.device)
    delta_g = gauss_encoder(I, g0)
    g = g0 + delta_g
    y_hat = gauss_cls(g)
    mu, quats, scale, op, colors = decode_gauss(g)
    I_hat = rasterize(mu, quats, scale, op, colors, **kwargs)
    return g, y_hat, I_hat

def clone_grads(params, grads):
    out = []
    for p, grad in zip(params, grads):
        out.append(torch.zeros_like(p) if grad is None else grad.detach().clone())
    return out

enc_params = [p for p in gauss_encoder.parameters() if p.requires_grad]
for step, (I, y) in enumerate(loader):
    g, y_hat, I_hat = forward_gvit(I)
    loss_pixel = bce(I_hat, I)
    loss_perc = dssim(I_hat, I)
    loss_rec = loss_pixel + plambda * loss_perc
    loss_cls = ce(y_hat, y)
    loss = loss_rec + clambda * loss_cls
    opt_enc.zero_grad(set_to_none=True)
    opt_cls.zero_grad(set_to_none=True)
    loss.backward()
    opt_enc.step()
    opt_cls.step()
    if step % guide_every == 0:
        g, y_hat, I_hat = forward_gvit(I)
        loss_pixel = bce(I_hat, I)
        loss_perc = dssim(I_hat, I)
        loss_rec = loss_pixel + plambda * loss_perc
        loss_cls = ce(y_hat, y)
        rec_grads = torch.autograd.grad(
            loss_rec,
            enc_params,
            retain_graph=True,
            allow_unused=True,
        )
        cls_grads = torch.autograd.grad(
            loss_cls,
            enc_params,
            allow_unused=True,
        )
        rec_grads = clone_grads(enc_params, rec_grads)
        cls_grads = clone_grads(enc_params, cls_grads)
        opt_enc.zero_grad(set_to_none=True)
        for p, g_rec, g_cls in zip(enc_params, rec_grads, cls_grads):
            p.grad = g_rec + gamma * g_cls
        opt_enc.step()
        gauss_cls.requires_grad_(True)

def decode_gauss(g):
    # g stores the 9D learnable 2D parameters:
    # [p(2), s(2), alpha(1), rgb(3), op(1)].
    # gsplat expects a 14D 3D backend format:
    # [mu(3), quats(4), scale(3), op(1), colors(3)].
    # We constrain that backend representation to 2D by fixing mu_z = 0,
    # using a fixed scale for z, and zeroing quaternion x/y components.
    p, scale_xy, angle, colors, op = split(g, [2, 2, 1, 3, 1])
    p = tanh(p)
    mu = concat([p, zeros_like(angle)], dim=-1)
    scale_xy = scale_factor * sigmoid(scale_xy)
    scale_z = fixed_z_scale * ones_like(angle)
    scale = concat([scale_xy, scale_z], dim=-1)
    quats = quat_from_z_angle(angle)
    quats[..., [1, 2]] = 0
    op = sigmoid(op)
    colors = sigmoid(colors)
    return mu, quats, scale, op, colors
\end{lstlisting}
\end{algorithm}

\clearpage
\clearpage

\end{document}